\pdfoutput=1

\documentclass[11pt]{article}

\usepackage[]{acl}
\usepackage{fdsymbol}

\usepackage{times}
\usepackage{color} %
\usepackage{latexsym}
\usepackage{makecell}
\usepackage{tablefootnote}
\usepackage[toc,page]{appendix}
\usepackage{pifont}%

\usepackage{enumitem}
\usepackage[T1]{fontenc}

\usepackage[utf8]{inputenc}

\usepackage{microtype}

\usepackage{hyperref}
\usepackage{booktabs}
\usepackage{graphicx}
\usepackage{makecell}
\usepackage{footmisc}

\usepackage[skins, breakable]{tcolorbox}

\usepackage{floatrow}
\usepackage{microtype}

\usepackage{caption}
\usepackage{xcolor}

\usepackage{subcaption}
\usepackage{array, makecell} %

\usepackage{pifont}%
\usepackage{graphics}

\usepackage{tabularx,colortbl}

\usepackage{tikz}
\usepackage{collcell}

\usepackage{etoolbox}

\newcolumntype{?}{!{\vrule width 1.5pt}}

\newtoggle{inTableHeader}%
\toggletrue{inTableHeader}%
\newcommand*{\StartTableHeader}{\global\toggletrue{inTableHeader}}%

\definecolor{darkgreen}{rgb}{0.01, 0.75, 0.24}
\definecolor{amber}{rgb}{1.0, 0.49, 0.0}
\let\OldTabular\tabular%
\let\OldEndTabular\endtabular%
\renewenvironment{tabular}{\StartTableHeader\OldTabular}{\OldEndTabular\StartTableHeader}%

\usepackage{amsmath}
\usepackage{amsfonts,bm}
\usepackage{xspace}

\newcommand{\red}[1]{\textcolor{red}{#1}}

\definecolor{mypink3}{cmyk}{0, 0.7808, 0.4429, 0.1412}

\makeatletter   
\newcommand{\sveryshortarrow}[1][3pt]{\mathrel{%
    \vcenter{\hbox{\rule[-.5\fontdimen8\scriptfont3]
               {\scriptratio\dimexpr#1\relax}{\fontdimen8\scriptfont3}}}%
   \mkern-4mu\hbox{\let\f@size\sf@size\usefont{U}{lasy}{m}{n}\symbol{41}}}}
\makeatother

\def\eqref#1{equation~\ref{#1}}

\def\1{\bm{1}}

\def\m1{{\bm{1}}}

\DeclareMathAlphabet{\mathsfit}{\encodingdefault}{\sfdefault}{m}{sl}
\SetMathAlphabet{\mathsfit}{bold}{\encodingdefault}{\sfdefault}{bx}{n}

\usepackage[capitalize,noabbrev]{cleveref}
\Crefname{section}{\mbox{\S\hspace*{-0.25ex}}}{\mbox{\S\hspace*{-0.25ex}}}
\Crefname{equation}{Eq.}{Eqs.}
\Crefname{figure}{Fig.}{Figs.}
\Crefname{table}{Table}{Tables}
\Crefname{appendix}{\S$\!$}{\S$\!$}

\usepackage[colorinlistoftodos,prependcaption,textsize=tiny]{todonotes}

\usepackage{soul}

\usepackage{float}
\definecolor{azure}{rgb}{0.0, 0.5, 1.0}
\definecolor{darkbrown}{rgb}{0.4, 0.26, 0.13}
\usepackage{multirow}%
\usepackage{hhline}%

\newcommand{\model}{{ChartGemma}}

\title{{\model}: Visual Instruction-tuning for Chart Reasoning in the Wild}

\author{
Ahmed Masry$^{\clubsuit}$\thanks{\ \ $^{\dagger}$Equal contribution.} \quad Megh Thakkar$^{\varheartsuit}$\footnotemark[1] \quad Aayush Bajaj$^{\varheartsuit\dagger}$ \quad Aaryaman Kartha$^{\clubsuit\dagger}$\\ \bf Enamul Hoque$^{\clubsuit}$ \quad Shafiq Joty$^{\vardiamondsuit\spadesuit}$\\
$^\clubsuit$York University, Canada \quad 
$^\varheartsuit$MILA - Quebec AI Institute \\
$^\vardiamondsuit$Salesforce Research \quad $^\spadesuit$Nanyang Technological University, Singapore
\\
\{masry20, aarykary, enamulh\}@yorku.ca \\
\{megh.thakkar, aayush.bajaj\}@mila.quebec, sjoty@salesforce.com
}

\begin{document}
\maketitle

\begin{abstract}

Given the ubiquity of charts as a data analysis, visualization, and decision-making tool across industries and sciences, there has been a growing interest in developing pre-trained foundation models as well as general purpose instruction-tuned models for chart understanding and reasoning. However, existing methods suffer crucial drawbacks across two critical axes affecting the performance of chart representation models: they are trained on data generated from underlying data tables of the charts, ignoring the visual trends and patterns in chart images, \emph{and} use weakly aligned vision-language backbone models for domain-specific training, limiting their generalizability when encountering charts in the wild. We address these important drawbacks and introduce ChartGemma, a novel chart understanding and reasoning model developed over PaliGemma. Rather than relying on underlying data tables, ChartGemma is trained on instruction-tuning data generated directly from chart images, thus capturing both high-level trends and low-level visual information from a diverse set of charts. Our simple approach achieves state-of-the-art results across $5$ benchmarks spanning chart summarization, question answering, and fact-checking, and our elaborate qualitative studies on real-world charts show that ChartGemma generates more realistic and factually correct summaries compared to its contemporaries. We release the code, model checkpoints, dataset, and demos at \href{https://github.com/vis-nlp/ChartGemma}{https://github.com/vis-nlp/ChartGemma}.

\end{abstract}

\section{Introduction}

Language-augmented vision foundation models or vision-language models (VLMs) have proven to be effective in tackling numerous real-world multimodal tasks such as visual segmentation, captioning, question answering, and generation and editing~\cite{li2023llava, zhu2023minigpt}. Though these models excel when used for general purpose applications in the wild, they often fail to tackle tasks that require specialized understanding and decoding of patterns and visualizations~\cite{han2023chartllama}. An important domain-specific usage of VLMs is for understanding and reasoning over charts, given their ubiquity as a data analysis, visualization, and decision-making tool across businesses, economies, and scientific fields~\cite{hoque2022chartSurvey}. This has naturally led to the development of more specialized foundation models pre-trained on massive amounts of structured and often chart-specific data~\cite{liu2022matcha, masry2023unichart}. These models are, however, trained on a limited source of resources and focus on a specific set of tasks, constraining their real-world applicability~\cite{masry2024chartinstruct}.

Developing over the success of instruction-tuning enabling models to generalize to more tasks and applications~\cite{instructgpt}, there have been attempts at 'instruction-tuning' VLMs to endow them the ability to understand charts in more realistic and fundamental settings~\cite{meng2024chartassisstant}. These approaches generally depend on two crucial factors impacting their effectiveness: (i) Instruction-tuning dataset -- these methods either use the underlying data tables from existing web sources~\cite{masry2024chartinstruct} or use synthetically generated data-tables~\cite{han2023chartllama} from LLMs such as GPT-4~\cite{openai2023gpt4} to curate the instruction-tuning data, and (ii) Base model -- the existing methods either use chart-specific pre-trained models like UniChart~\cite{masry2023unichart} or VLMs pre-trained with weak image-text alignment such as LLaVA~\cite{li2023llava}. However, in existing methods, both these factors have critical drawbacks impacting their ability to understand real-world complex charts.

\begin{table*}[t]
\centering
\scalebox{0.50}{
\begin{tabular}{p{0.5\linewidth}p{0.6\linewidth}p{0.7\linewidth}}
\Xhline{4\arrayrulewidth}
\textbf{Chart Image} & \textbf{Summary Generated from the Data Table} & \textbf{Summary Generated from Chart Image}  \\ \hline \hline
\multicolumn{1}{p{0.5\linewidth}|}{\raisebox{-\height}{\includegraphics[width=0.5\textwidth]{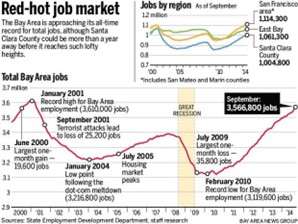}}} &
The provided data table shows the \textcolor{darkgreen}{total number of jobs in the Bay Area, as well as the number of jobs in San Francisco, the East Bay, and Santa Clara County}, from 2000 to \textcolor{red}{2026}. \textcolor{red}{Unfortunately, the table only provides data for the year 2000, with the remaining years missing data for all regions except the total Bay Area jobs.}
\textcolor{red}{While we can see the total Bay Area job growth from 3.3 million in 2000 to a projected 4.0 million in 2026, we lack the specific data to assess job trends within San Francisco, East Bay, and Santa Clara County over the years.}
To draw meaningful conclusions about job market trends in these regions, it is necessary to have complete data for all years and all regions.
 & The chart shows the job market in the Bay Area from 2000 to \textcolor{red}{2012}. \textcolor{darkgreen}{The Bay Area is approaching its all-time record for total jobs, although Santa Clara County could be more than a year away before it reaches such lofty heights}. \textcolor{darkgreen}{The chart shows that the Bay Area job market experienced a significant drop during the Great Recession, but has since recovered and is now at an all-time high.} \textcolor{darkgreen}{The chart also shows the number of jobs in different regions of the Bay Area, including San Francisco, East Bay, and Santa Clara County.} The chart is a good indicator of the overall health of the Bay Area economy.
 \\ \midrule
\multicolumn{1}{p{0.5\linewidth}|}{\raisebox{-\height}{\includegraphics[width=0.5\textwidth]{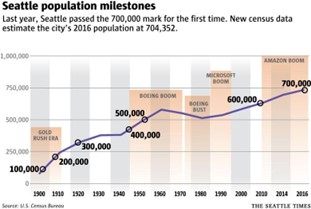}}} &
The data table shows the population of a city or region from 1900 to 2016. The population grew steadily from 100,000 in 1900 to 550,000 in 1960. After a slight decline in the 1970s, the population rebounded and reached 704,352 in 2016. 

& The chart shows the population milestones of Seattle from 1900 to 2016. \textcolor{darkgreen}{The population has been steadily increasing over the years, with significant booms during the Gold Rush Era, the Boeing Boom, the Microsoft Boom, and the Amazon Boom. The population reached 700,000 for the first time in 2015. The population peaked at around 600,000 during the Boeing Boom in the 1960s, before declining slightly during the Boeing Bust in the 1970s. The population increased again during the Microsoft Boom in the 1990s and has been steadily increasing ever since.} The population of Seattle reached 704,352 in 2016. The data is based on U.S. Census Bureau data.

\\ \midrule
\multicolumn{1}{p{0.5\linewidth}|}{\raisebox{-\height}{\includegraphics[width=0.5\textwidth]{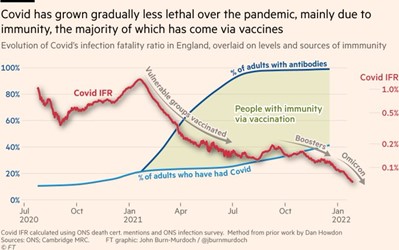}}} &
The data table shows the trend of Covid-19 related metrics from July 2020 to January 2022. It reveals a decline in the Infection Fatality Rate (IFR) from 80\% in July 2020 to 25\% in January 2022. This coincides with a rise in the percentage of adults who have had Covid, increasing from 10\% in July 2020 to 60\% in January 2022. The percentage of adults with antibodies also shows a significant increase, going from 0\% in July 2020 to 100\% in January 2022. These trends suggest a possible correlation between increased exposure to Covid-19 and a lower IFR, likely due to herd immunity and improved treatment options. 

& This chart shows the evolution of \textcolor{darkgreen}{Covid's infection fatality ratio in England, overlaid on levels and sources of immunity.} The chart shows that the Covid infection fatality ratio has decreased significantly since the beginning of the pandemic. This is likely due to the increasing levels of immunity in the population, both from vaccination and from infection. The chart also shows that the Covid infection fatality ratio has been declining at a faster rate since the introduction of vaccines. \textcolor{darkgreen}{It also shows that the most vulnerable groups have been vaccinated and that this has had a significant impact on reducing the number of deaths from Covid.} \textcolor{darkgreen}{The infection fatality ratio has also decreased with the introduction of boosters.} This demonstrates the impact of the vaccine in helping to protect people from serious illness and death from Covid.

\\ \hline
\end{tabular}
}
\caption{\label{tab:comparison}Summaries generated from the same LLM, Gemini Flash 1.5, when using the data table and the chart image, highlighting the importance of understanding the visual attributes to generate more appropriate chart instructions. 
}
\vskip -1ex
\end{table*}

Existing methods are restricted to charts that either have an underlying data table or require methods to extract them from the charts, often with low accuracy which are used for instruction-tuning data generation. These data tables are often incapable of capturing numerous nuanced details in the complex charts used in real-world applications (\Cref{tab:comparison}). Also, in many scenarios, we are concerned with representing or understanding general trends in the charts and not individual data points. On the model side, existing methods use backbones in which the vision encoder and LLM are weakly-aligned, either due to limited data or architecture, limiting their generalizability to represent real-world charts. Instruction-tuning a strongly aligned base VLM can capture the intricacies among diverse chart elements and corresponding text more efficiently. We hypothesize that formulating a simple approach addressing these drawbacks can lead to an effective foundation model capable of complex chart understanding and reasoning in the wild. 

We propose ChartGemma, an instruction-tuned multimodal model for chart understanding and reasoning. ChartGemma uses instruction-tuning data for chart representation learning that is directly generated from the chart images, capturing more diverse and relevant information while preserving complex visual features. This also enables us to utilize a much broader array of charts available across the web as we are not restricted by the availability of underlying data tables. ChartGemma develops over PaliGemma~\cite{beyer2024paligemma} which has been trained on a much larger alignment dataset. Since ChartGemma uses PaliGemma as its backbone, it is also much smaller than existing chart understanding models, making it suitable for real-world applications. We evaluate ChartGemma across 5 benchmarks spanning chart summarization, question answering, and fact-checking, obtaining state-of-the-art results compared to existing methods. Our qualitative studies also demonstrate that ChartGemma produces more faithful and realistic summaries of complex charts as compared to other methods. Through our elaborate analysis, we put forward ChartGemma as an effective model capable of understanding and reasoning over real-world charts. Our main contributions are:

\begin{figure*}[t!]
     \centering
        \includegraphics[width=.98\textwidth]{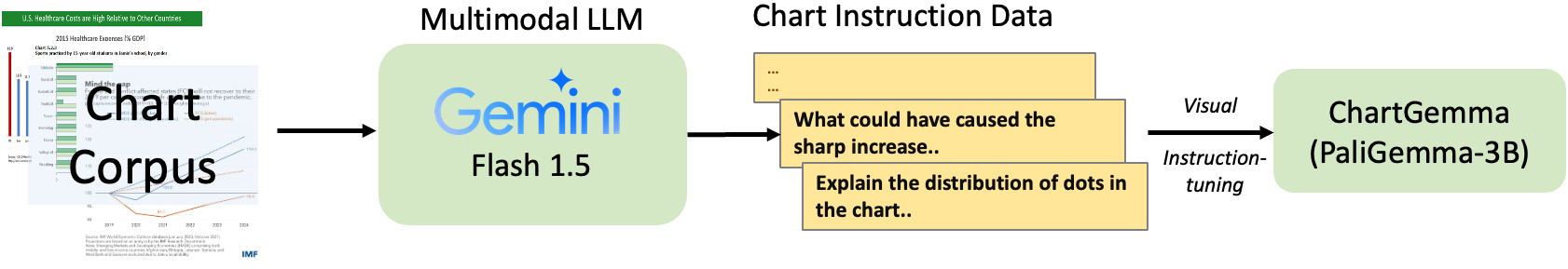}
         \caption{\small{
         The instruction-tuning data generation process. Chart images are input into Gemini Flash 1.5, which generates visual chart instructions used to fine-tune our model, ChartGemma (please refer to \cref{sec:main_pipeline}).
         }
         \vspace{-2mm}
         }
    \label{fig:instruct-tuning-process}
\end{figure*}

\begin{itemize}[leftmargin=12pt]
\itemsep0em
    \item We present ChartGemma, a first-of-its-kind multimodal model instruction-tuned for chart understanding and reasoning using data directly generated from chart images.
    \item ChartGemma utilizes a stronger backbone model and more representative instruction-tuning data, rendering it effective in tackling existing benchmarks across chart summarization, question answering, and fact-checking while being significantly smaller than its counterparts.
    \item Our extensive quantitative and qualitative studies reveal that ChartGemma generates more faithful and %
    human-like
    summaries and is extremely capable in understanding and representing complex real-world charts in the wild.
\end{itemize}

\section{Chart Instruction Data Generation}
\label{sec:main_pipeline}

In this section, we present the details of generating our instruction-tuning dataset. We start by curating a diverse corpus of charts that encompasses a range of visual styles and elements (\cref{subsec:chart_corpora_collection}), and then use it to generate the visual instruction-tuning data directly from the charts (\cref{subsec:visual_chart_instruction}). We illustrate our data generation pipeline in \cref{fig:instruct-tuning-process}.

\subsection{Assembling the Chart Corpus}
\label{subsec:chart_corpora_collection}

Our chart corpus is assembled using a combination of various sources across three categories: (i) Synthetically generated charts from sources such as PlotQA~\cite{plotqa}, (ii) Curated charts from specialized websites such as Statista which typically exhibit limited visual diversity, and (iii) In-the-wild charts harvested from the broader web, such as WebCharts~\cite{masry2024chartinstruct}, noted for their extensive stylistic variety. While prior approaches used accompanying metadata (e.g., titles, data tables, annotations) to generate instructions from LLMs~\cite{han2023chartllama, meng2024chartassisstant}, our method exclusively utilizes the chart images themselves for generating instruction-tuning data. This approach also allows us to bypass the constraints imposed by metadata availability. In total, our corpus consists of 122,857 chart images. We provide an elaborate breakdown of the chart source and the statistics across each category in \cref{tab:instructiondata}.

\subsection{Visual Chart Instructions}
\label{subsec:visual_chart_instruction}

We use chart images directly from the above assembled corpus to generate visual instruction-tuning data. This enables us to synthesize data that can train a model to capture not just point information, but complex trends and relations among the chart elements.
Following \citet{masry2024chartinstruct}, we generate data across two categories: (i) predefined tasks, which align with common real-world scenarios and benchmarks, and (ii) open-ended tasks. For pre-defined tasks, we generate data for,

\noindent \textbf{1. Chain-of-thought (CoT)} involves prompting the model with complex reasoning questions and enhances the visual reasoning capabilities of the model by guiding it through the problem-solving process in a structured manner.

\noindent \textbf{2. Summarization} involves prompting the model to generate summaries that succinctly capture the key insights and trends from a chart image that effectively communicates the primary data narratives. 

\noindent \textbf{3. Fact Checking} asks the model to determine whether stated facts are supported or refuted by the data presented in a chart image. Alongside data generated from our corpus, we use the training sets of existing chart fact-checking tasks~\cite{akhtar2023reading, akhtar2023chartcheck} in our instruction-tuning data. 

\noindent \textbf{4. Chart-to-Markdown} tasks the model with generating the underlying data tables from a chart image in Markdown format. This approach simplifies rendering and parsing the tables, enhancing their accessibility and usability.

\noindent \textbf{5. Program Aided Design}~\cite{gao2022pal} requires the model to generate executable code that performs necessary calculations and outputs the final answer, delegating complex and challenging mathematical operations to the code interpreter. Alongside synthetic data generated from our corpus, we use the Multimodal LLM to create executable codes for questions in the training split of the ChartQA dataset \cite{MSMmasry2022chartqa}, augmenting our instruction-tuning data with human-written questions and their corresponding code.

\noindent \textbf{Open-ended Tasks} We enrich our instruction-tuning data by prompting the Multimodal LLM to generate a variety of tasks typical in real-world scenarios. This approach enhances the generalizability of our models and extends their applicability to diverse real-world settings. 
Example open-ended tasks include justifying temporal or time-series based trends observed in the chart, describing the different visual elements such as lines, colors, and legends represented by the chart, critically analyzing and comparing visual information, etc. We present concrete examples in \cref{appendix:data_analysis}.

\noindent We use Gemini Flash-1.5 \cite{geminiteam2023gemini} due to its robust multimodal performance, cost-effectiveness, and high API rate limits.

\subsection{Key Dataset Characteristics}
\label{subsec:dataset_characteristics}

To underscore the distinct innovations of our dataset relative to prior works, we examine two critical elements: the visual attributes and the quality of the chart instructions.

\textbf{Visual Attributes} Our instruction-tuning dataset features a wide range of instructions that emphasize the visual attributes of chart images. As illustrated in \cref{fig:visual_examples} in Appendix \ref{appendix:data_analysis}, the examples highlight various visual elements such as lines, shapes, colors, trends, chart types, and positions, all of which are frequently referenced in real-world scenarios. These enhance the model's visual reasoning capabilities, enabling real-world applications.

\textbf{Quality} To demonstrate the strength of our approach in generating high-quality and accurate instructions, we evaluated 100 randomly sampled synthesized instructions. We found that our instructions accurately reflected the chart content in 82\% of the cases, which is a significant improvement over the 61\% accuracy reported for the ChartInstruct dataset~\cite{masry2024chartinstruct}. Additionally, we observed 8\% partially correct answers, similar to that as reported by ChartInstruct. We attribute this improvement in quality to our method’s reliance on the chart images, rather than using automatically generated and often erroneous data tables.

\section{Modeling and Methodology}

\begin{figure}[t!]
     \centering
        \includegraphics[width=1\textwidth]{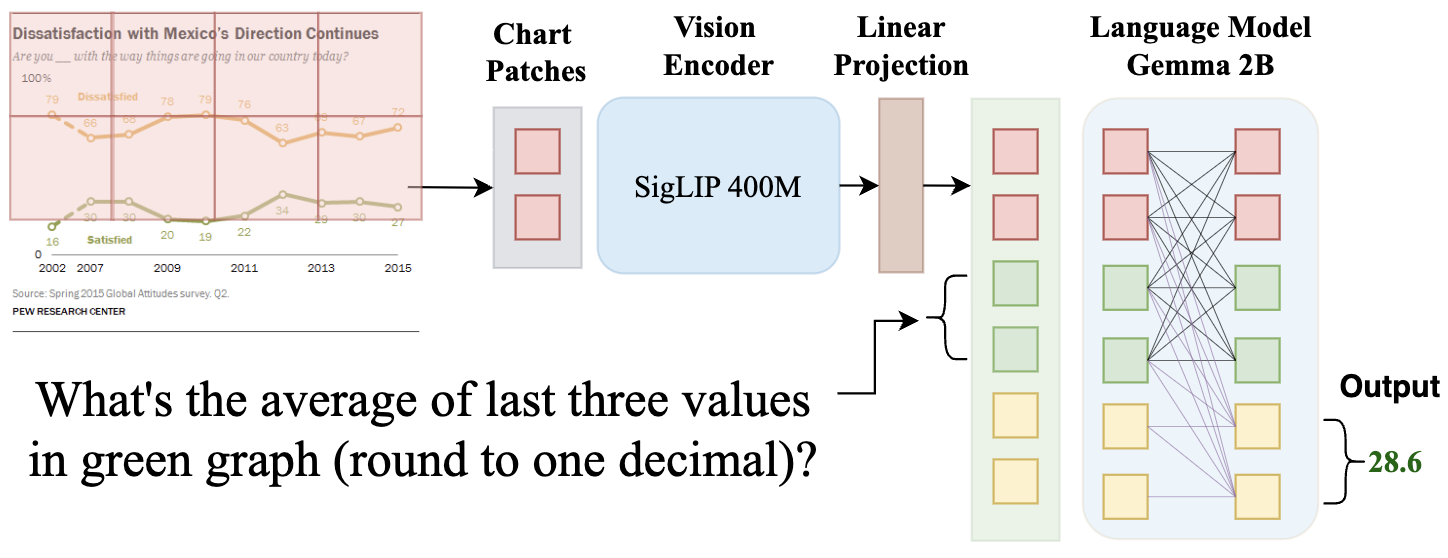}
        \vspace{-3mm}
         \caption{\small{
         ChartGemma architecture featuring the SigLIP vision encoder and the Gemma-2B language model. Visual tokens are depicted in red, prefix tokens in green, and suffix tokens in yellow. Full attention is applied between visual and prefix tokens (indicated by black lines), while causal attention is used for suffix tokens (indicated by purple lines) which are generated autoregressively.}
         }\label{fig:end-to-end-both}
         \vspace{-4mm}
\end{figure}

\subsection{Architecture}

\model{} uses PaliGemma~\cite{beyer2024paligemma} as the backbone architecture, which comprises of the following two components:

\paragraph{Vision Encoder: SigLIP}~\cite{siglip} is a vision transformer (ViT) encoder
. Unlike CLIP-like ViTs~\cite{clipvisual} which use 
contrastive loss 
on large batches of image-text pairs, 
SigLIP is trained on single image-text pairs independently as a binary classification task.

\paragraph{Language Model: Gemma-2B}~\cite{gemma} is decoder-only transformer-based~\cite{vaswani2017attention} LLM trained on 3 trillion tokens with a context length of 8,196 tokens. Its pretraining data mainly consists of English documents, maths, and code, making it suitable for chart understanding tasks requiring strong reasoning capabilities.

We present ChartGemma's architecture in \cref{fig:end-to-end-both}. The input image is taken in 448x448 resolution and divided into 14x14 pixel patches, each of which is fed into the vision encoder as a separate token. The outputs from the vision encoder are passed through a linear layer that maps the visual features into the LLM embedding space. These visual tokens are then concatenated with the input text embeddings and passed to Gemma-2B. Unlike previous VLLMs~\cite{li2023llava} that indiscriminately apply a causal mask on all image and text tokens, Gemma-2B applies full attention over the input visual and text tokens while a causal mask is applied on the output tokens. This improves the contextual understanding of the image particularly for representing complex relationships among objects. We believe this property provides further advantages when learning representations for chart images containing numerous nuanced complexities.

\begin{table*}[t!]
 \small
 \centering
\resizebox{\linewidth}{!}{
 \scalebox{0.50}{
 \begin{tabular}{l|c|cccccc}
  
  \toprule
  \multirow{3}{*}{} & \multirow{3}{*}{} & \multicolumn{3}{c}{ChartQA} & \multicolumn{3}{c}{Chart Fact Checking} \\ 
  \multirow{3}{*}{} & \multirow{3}{*}{} & \multicolumn{3}{c}{\small ($Relaxed\ Accuracy$)} & \multicolumn{3}{c}{\small ($Accuracy$)}  \\
  \cmidrule(lr){3-5} \cmidrule(lr){6-8} 
  
   Model & \#Params & aug. & human & avg. & ChartFC & ChartCheck T1 &  ChartCheck T2 \\

  \midrule
  \multicolumn{8}{l}{\textbf{Specialist Chart Models}}
  \\

  ChartBERT \cite{akhtar2023reading} & - & - & - & - & \underline{63.8} & - & -
  \\

  Pix2Struct \cite{lee2022pix2struct} & 282M & 81.6 & 30.5 & 56.0 & - & - & -
  \\
  
  Matcha\cite{liu2022matcha}  & 282M & \underline{90.2} & 38.2 & 64.2 & - & 62.80 & 61.40
  \\ 

  UniChart \cite{masry2023unichart}  & 201M & 88.56 & \underline{43.92} & \underline{66.24} & - & - & - \\

  \midrule

   \multicolumn{8}{l}{\textbf{\textcolor{gray}{Closed VLMMs}}} \\

  \textcolor{gray}{Gemini Pro} \cite{geminiteam2023gemini} & \textcolor{gray}{-} & \textcolor{gray}{-} & \textcolor{gray}{-} & \textcolor{gray}{74.1} & \textcolor{gray}{65.8} & \textcolor{gray}{-} & -
\\

  \textcolor{gray}{GPT4-V} \cite{openai2023gpt4}  & \textcolor{gray}{-} & \textcolor{gray}{-} & \textcolor{gray}{-} & \textcolor{gray}{78.5} & \textcolor{gray}{69.6} & \textcolor{gray}{-} & -
  \\ 

  \midrule

  \multicolumn{8}{l}{\textbf{Chart VLLMs}} \\

  ChartLlama \cite{han2023chartllama}  & 13B & 90.36 & 48.96 & 69.66 & - & - & -
  \\ 

  ChartAssisstant \cite{meng2024chartassisstant}  & 13B & \underline{93.90} & 65.90 & 79.90 & - & - & -
 \\ 

  ChartInstruct-Llama2 \cite{masry2024chartinstruct}  & 7B & 87.76 & 45.52 & 66.64 & 69.57 & 70.11 & 68.80
  \\ 
  
  ChartInstruct-Flan-T5-XL \cite{masry2024chartinstruct}  & 3B & 85.04 & 43.36 & 64.20 & 70.27 & \underline{72.03} & 73.80
  \\

  ChartGemma (Ours) & 3B & 90.80 & \underline{69.52} & \underline{80.16} & \underline{70.33} & 71.50 & \underline{74.31}
  \\
  
  \bottomrule
 \end{tabular}
 }
 }
 \caption{\small {Performance on closed-ended generation benchmarks: ChartQA, ChartFC, and ChartCheck. ChartGemma generally outperforms or matches the performance of all the baselines, while being significantly smaller than them (refer to \cref{subsec:closed_ended_performance}).
 }
 }
 \label{tab:results_new}
\end{table*}

\subsection{Training Setup}

Existing chart VLLMs~\cite{meng2024chartassisstant} typically employ a two-stage training approach that requires an initial step to align the vision encoder and the LLM for understanding chart features, followed by instruction-tuning. In contrast, we only use a single-stage approach where we directly finetune the backbone model on our instruction-tuning data. We believe that the first stage is required by current methods as the VLLM backbones are aligned using a limited amount of image-text pairs with restricted styles and diversity. In contrast, our backbone, PaliGemma, has been trained end-to-end on 10 billion image-text pairs covering a wide variety of styles. This makes our model more adaptable and generalizable to different real-world images (e.g., charts, infographics, documents). We freeze the vision encoder and only finetune the LLM during instruction-tuning. This helps in reducing the computational complexity and also improves training stability given the small batch size used for instruction-tuning PaliGemma.

\section{Experiments, Results, and Analyses}

\subsection{Experimental Setup}
\label{subsec:setup}

\paragraph{Baselines} \label{subsec:baselines}
We compare \model\ against baselines comprising of open-source chart-specialist models and VLLMs instruction-tuned on chart data, as well as state-of-the-art closed source multimodal LLMs. Chart-specialist models include ~\emph{ChartBERT} \cite{akhtar2023chartcheck}, ~\emph{Pix2Struct} \cite{lee2022pix2struct}, \emph{MatCha} \cite{liu2022matcha}, and \emph{UniChart}~\cite{masry2023unichart}. Chart VLLMs include \emph{ChartLlaMA}~\cite{han2023chartllama}, \emph{ChartAssistant}~\cite{meng2024chartassisstant}, and \emph{ChartInstruct}'s~\cite{masry2024chartinstruct} two variants with LLaMA2 and Flan-T5-XL. We also compare \model\ against two closed-source multimodal LLMs, namely Gemini Pro~\cite{geminiteam2023gemini} and GPT4-V~\cite{openai2023gpt4}. 

\paragraph{Downstream Tasks}
We evaluate ChartGemma on a diverse set of 5 established benchmarks evaluating chart representation and reasoning abilities: (i) ChartQA \cite{MSMmasry2022chartqa} -- a factoid chart question answering dataset, (ii) ChartFC \cite{akhtar2023reading} and (iii) ChartCheck~\cite{akhtar-etal-2023-reading} -- chart fact checking datasets, (iv) OpenCQA \cite{open-CQA} -- an open-ended chart question answering dataset, and (v) Chart2Text \cite{chart-to-text-acl} -- a chart summarization dataset. While ChartQA and ChartFC focus on closed-ended generation, OpenCQA and Chart2Text evaluate open-ended generation abilities of the models.
We also manually curate a set of $100$ charts downloaded from the web completely unseen by any model. We refer to this set as 'Web' in our results, and use them for comparing the summarization ability of the models.

\paragraph{Evaluation Metrics} Following existing works, we use relaxed accuracy (RA) for ChartQA, accuracy for ChartFC, and use GPT4 as a judge for open-ended generation tasks, i.e. Chart2Text, OpenCQA, and our curated Web set of charts and measure the informativeness and factual correctness on a scale of 1-5~\cite{post-2018-call}. 

To ensure the reproducibility of our work, we present the hyperparameters of our instruction-tuning and downstream task experiments in \cref{subsec:appendix-hyperparameters}. All experiments were conducted on a 4 A100 GPUs (80GB) machine using the JAX framework\footnote{https://github.com/google/jax}.

\subsection{Performance on closed-ended tasks}
\label{subsec:closed_ended_performance}

We compare the performance of ChartGemma to the various baselines on the closed-ended tasks, namely ChartQA and ChartFC, and present the results in Table \ref{tab:results_new}. We see that Chart VLLMs are generally the better performing set of models compared to specialist chart models. Within Chart VLLMs, we observe that ChartGemma performs the best on ChartQA in terms of the average overall performance and on both the synthetic ChartFC and real-world-based ChartCheck test splits. Particularly, the performance improvements on ChartCheck when using ChartGemma, which is a zero-shot evaluation, can be attributed to the fact that our instruction-tuning dataset is specifically designed to generalize to more realistic charts encountered in this particular evaluation. We observe that it is also powerful for its small size of 3 billion parameters, and only lags in performance to the 13 billion parameter ChartAssistant on the augmented set of ChartQA. The significant improvement of ChartGemma over ChartAssistant on the human-generated split of ChartQA indicates better generalization abilities in understanding more realistic instructions for complex charts.

Given the state-of-the-art performance of ChartGemma, we next perform a series of ablations to test our hypothesis on the criticality of having (i) an instruction-tuning dataset derived from chart images rather than the underlying data tables, and (ii) the importance of a strong backbone model.

\begin{table}[t!]
 \centering
\resizebox{\linewidth}{!}{
 \scalebox{0.5}{
 \begin{tabular}{l|cccccc}
  
  \toprule
  \multirow{3}{*}{} & \multicolumn{3}{c}{ChartQA} & \multicolumn{3}{c}{Chart Fact Checking} \\ 
  \multirow{3}{*}{} & \multicolumn{3}{c}{\small ($Relaxed\ Accuracy$)} & \multicolumn{3}{c}{\small ($Accuracy$)}  \\
  \cmidrule(lr){2-4} \cmidrule(lr){5-7} 
  
   Model & aug. & human & avg. & ChartFC & ChartCheck T1 &  ChartCheck T2 \\

   \midrule
   
PaliGemma  & - & - & 71.36 & 58.26 & 67.34 & 68.50
  \\
  
  PaliGemma+ChartInstruct & 70.24 & 33.84 & 52.04 & 48.58 & 54.21 & 51.78
  \\

  LLaVA+Our dataset & 61.12 & 51.12 & 56.12 & 61.28 & 70.22 & 70.03
  \\

  ChartGemma (Ours) & 89.44 & 64.80 & 77.12 & 69.95 & 72.03 & 73.80
  \\
  
  \bottomrule
 \end{tabular}
 }
 }
 \vspace{-3mm}
 \caption{\small {Ablation results validating our hypothesis on the effect of our instruction-tuning data and backbone model on downstream tasks (refer to \cref{subsec:closed_ended_performance}).
 }
 \vspace{-3mm}
 }
 \label{tab:results_ablations_instruct}
\end{table}

\paragraph{Effect of the instruction-tuning data} To validate the effectiveness of synthesizing instruction-tuning data directly using the chart images as compared to using their underlying data tables, we compare ChartGemma with a version of PaliGemma instruction-tuned on the dataset presented in ChartInstruct~\cite{masry2024chartinstruct},  which was generated using the chart data tables. We present the results in \cref{tab:results_ablations_instruct}. We observe remarkable improvements when using our instruction-tuning data compared to the data proposed by ChartInstruct. The improvements are stark on the human split of ChartQA, indicating that ChartGemma is very efficient in following real-world human instructions. The significantly weak performance of ChartGemma when using the dataset from ChartInstruct is in-line with the observations of the author mentioning a low (61 \%) accuracy of the synthetically generated instruction-tuning data~\cite{masry2024chartinstruct}.

\paragraph{Effect of the backbone model} 
We probe the effect of using PaliGemma as the backbone model for ChartGemma, which has better image-text alignment compared to other VLMs, on the downstream performance. We follow existing works~\cite{han2023chartllama, masry2024chartinstruct} that use LLaVA~\cite{liu2023visual} as a backbone and train LLaVA-1 with our instruction-tuning data.  We compare this variant (LLaVA+Our dataset) with ChartGemma in \cref{tab:results_ablations_instruct} and observe that ChartGemma performs significantly better as compared to using LLaVA as our backbone. This validates our hypothesis that initializing our architecture with a strongly aligned model leads to better char understanding, reasoning, and generalization capabilities.

\subsection{Performance on open-ended tasks}
\label{subsec:open-ended-eval}

\begin{figure}[t!]
    \centering
    \includegraphics[width=\linewidth]{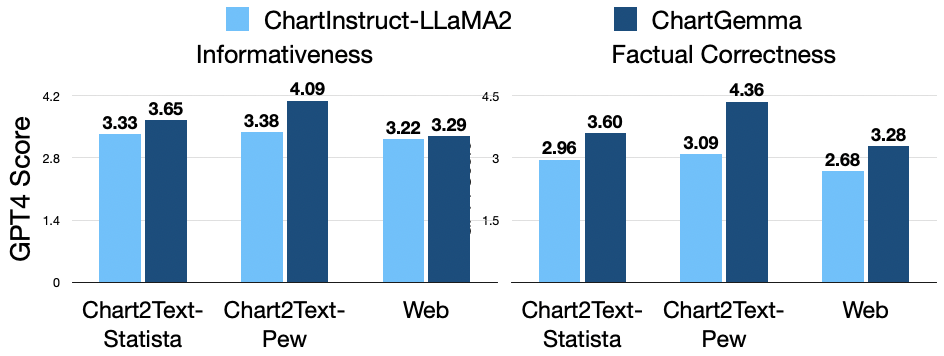}
    \caption{GPT4 scores (from 1-5, with 5 being the highest) on the informativeness and factual correctness of outputs generated by ChartInstruct-LLaMA2 and ChartGemma (refer to \cref{subsec:open-ended-eval}).}
    \label{fig:gpt4-eval}
    \vskip -1ex
\end{figure}

We next compare the performance of ChartGemma with the baselines on chart understanding and reasoning based open-ended generation benchmarks, OpenCQA~\cite{open-CQA}, Chart2Text~\cite{chart-to-text-acl}, and our curated 'Web' set. We do not use the BLEU~\cite{papineni-etal-2002-bleu} scores for comparison as done by previous works, due to the numerous criticisms of it as an indicative metric~\cite{callison-burch-etal-2006-evaluating, smith-etal-2016-climbing} and follow the widespread practice of using strong LLMs as a judge due to their high agreement with human annotators~\cite{zheng2023judging}. We use GPT4 to evaluate the informativeness and factual correctness of the outputs generated by the models and present the scores in \cref{fig:gpt4-eval}\footnote{We show the extended results in Appendix \ref{sec:appendix-gpt4-eval}.}. We see that the outputs generated by ChartGemma are generally scored higher as compared to ChartInstruct. We particularly see significant improvement in the factual correctness of the outputs of ChartGemma, probably due to the fact that our instruction-tuning data synthesized using the chart images captures more complex visual elements and PaliGemma being strongly aligned leads to better understanding and reasoning over the charts. Our findings overall indicate that ChartGemma is able to produce more informative outputs while also being factually correct in terms of long-form answering or summarization for the charts.

\subsection{Human Evaluation on Summarization}
\label{subsec:human_eval}

\begin{figure}[t!]
    \centering
    \includegraphics[width=\linewidth]{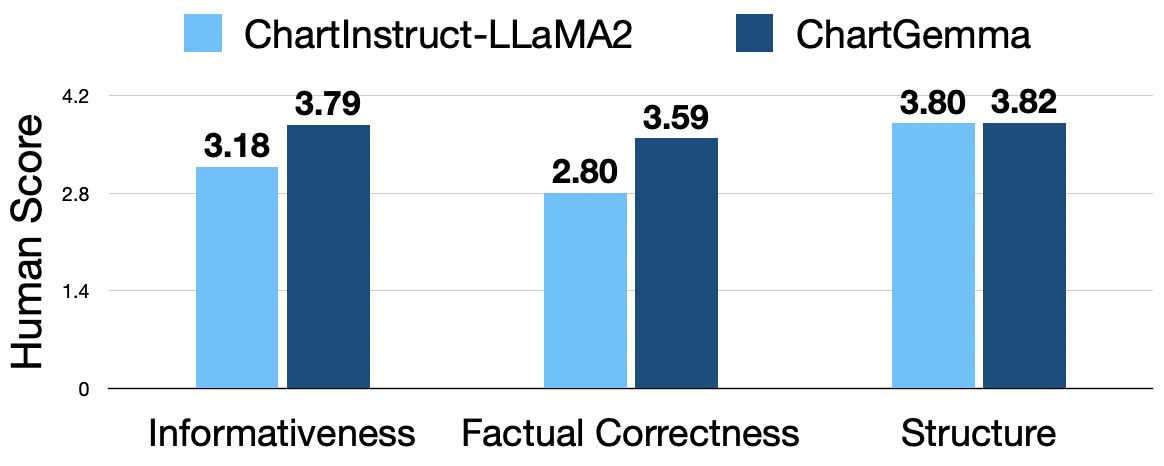}
    \caption{Human evaluation scores on the informativeness, factual correctness, and structure of outputs generated by ChartInstruct-LLaMA2 and ChartGemma.}
    \label{fig:human-eval}
\end{figure}

Though using online LLMs like GPT4 as a judge has been shown to have a high correlation with human annotation~\cite{zheng2023judging}, there haven't been studies on measuring this correlation explicitly for chart understanding tasks. Hence, to ensure our observations, evaluations, and conclusions are robust, we perform a human study on the manually curated set of 100 charts, 'Web'. Similar to GPT4 evaluation, we compare the informativeness, factual correctness, and structure of the outputs generated by ChartGemma with ChartInstruct-LLaMA2.

We first use ChartInstruct-LLaMA2 and ChartGemma to generate summaries for these samples in the Web set. We then ask 2 different annotators to rate all the responses based on the above metrics (informativeness, factual correctness, structure) from 1-5 (5 being the highest) so we can also measure agreement between the annotations\footnote{ We found a Cohen's Kappa of 0.538 for the agreement.}. We present the outputs randomly to the annotators to prevent any biases towards the models and present the evaluation results in \cref{fig:human-eval}.

From \cref{fig:human-eval}, we observe that ChartGemma consistently outperforms or matches ChartInstruct-LLaMA2 on all the metrics, and the findings are in-line with those observed when using GPT4 for evaluation (Section \ref{subsec:open-ended-eval}). 
We observe that ChartGemma is equally well structured, yet is more informative and significantly more factually correct. Better informativeness probably stems from the fact that ChartGemma is trained on data generated from the chart images and not just the underlying data tables, enabling it to learn high level trends and concepts specific to charts. Furthermore, our instruction-tuning data and a strong backbone model promote capturing more complex visual elements of charts, leading to more factual correctness.
Overall, since our evaluation is performed on charts sampled randomly in the wild from the web, ChartGemma's strong performance validates its effectiveness as a strong candidate in understanding and reasoning over real-world charts.

\subsection{Error Analysis and Challenges}

We analyzed the outputs of our model, ChartGemma, to understand the shortcomings and areas for improvement. We have discovered the following three patterns of errors.

\noindent{\bf High Resolution Charts}
Charts with very large, often skewed dimensions, present challenges for our model, which uses an input resolution of 448x448. Resizing these large images can cause written text to become unreadable, leading to errors in the predicted labels and numerical values, as depicted in \cref{fig:WrongSamples}. Although PaliGemma offers a variant supporting up to an 896x896 input resolution, it operates significantly slower than the 448x448 version, making it impractical for use on consumer-level machines and GPUs.

\noindent{\bf Coding Errors}
While ChartGemma demonstrated state-of-the-art performance on the ChartQA benchmark, excelling in complex numerical reasoning and compositional questions, it occasionally generates erroneous code that cannot be executed. As depicted in \cref{fig:WrongSamples}, the model sometimes refers to undeclared variables within the code. We believe that integrating an LLM with enhanced coding capabilities could further improve our performance on the ChartQA benchmark.

\noindent{\bf Charts with Complex Visual Styles }
Although our instruction-tuning corpus predominantly features real-world charts from the broad web, ChartGemma tends to exhibit lower factual correctness and informativeness when evaluated on these charts compared to those from specialized websites like Pew or Statista, which have less visual diversity. This disparity, illustrated in \cref{fig:gpt4-eval}, highlights the need for further enhancements to improve the generalizability of chart understanding models across various visual styles.

\subsection{Convergence of ChartGemma}
\label{subsec:convergence}

\begin{figure}[t!]
    \centering
    \includegraphics[width=0.75\linewidth]{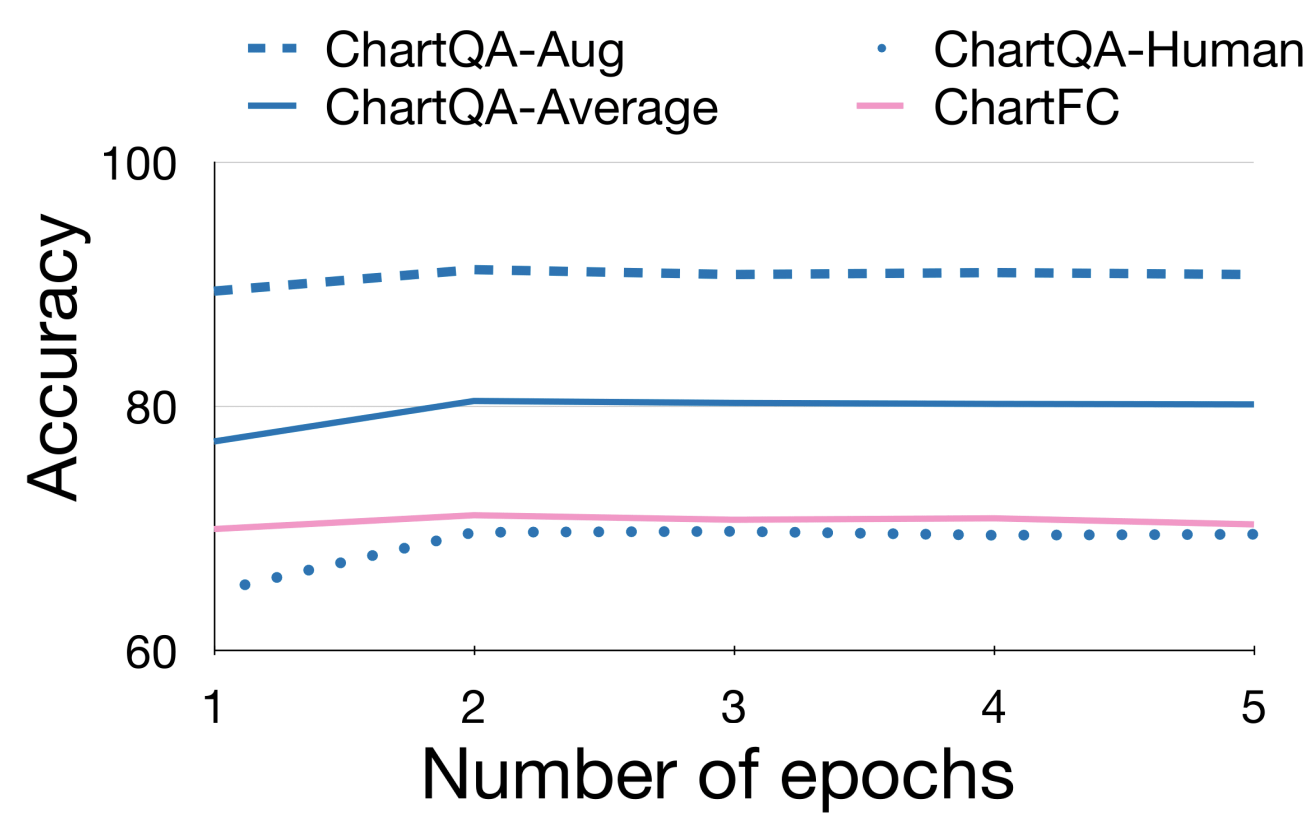}
    \caption{Effect of the number of epochs on instruction-tuning ChartGemma. We observe very quick convergence during training (refer to \cref{subsec:convergence}). For ChartQA, accuracy is relaxed accuracy (\cref{subsec:setup}).}
    \label{fig:epoch-ablation}
    \vskip -1ex
\end{figure}

We probe the learning dynamics of ChartGemma by checking the downstream accuracy with the number of instruction-tuning epochs and present the trends in \cref{fig:epoch-ablation}. We interestingly observe that ChartGemma converges very quickly, with the best performance observed at epoch $2$. We attribute this characteristic to the strong alignment of PaliGemma rendering it effective in adapting to our relatively generalizable instruction-tuning dataset. This indicates that PaliGemma is a very efficient backbone for visual instruction-tuning of chart data, and might generalize when trained with a much larger number of samples as well. We leave this exploration as future work.

\section{Related Work}

\paragraph{Chart Representation Learning}
Chart understanding models initially were either fine-tuned from language or vision-language models~\cite{MSMmasry2022chartqa, ahmed-workshop-2021, lee2022pix2struct}, or pre-trained using chart-specific learning objectives~\cite{masry2023unichart, liu2022matcha}. 
Recently, instruction-tuning of pre-trained VLMs has been explored for enhancing the general applicability to charts~\cite{meng2024chartassisstant, han2023chartllama, masry2024chartinstruct, liu2023mmc}. Though these methods use diverse sources across the web and synthetic charts for generating instruction-tuning data, they utilize the underlying data table of the charts and train a weakly-aligned backbone VLM.

\paragraph{Chart Modeling Benchmarks}
With charts being the standard medium for data visualization and data-driven decision making, diverse benchmarks have been proposed to evaluate the abilities of LLMs and VLMs on chart understanding. These benchmarks range from close-ended tasks such as question answering~\cite{plotqa,masry-etal-2022-chartqa} to open-ended generation such as explanation generation in OpenCQA~\cite{open-CQA} and summarization~\cite{chart-to-text-acl}. Chart-specific benchmarks evaluate the ability of models to convert charts into data tables~\cite{Choi2019VisualizingFT, masry2023unichart} or evaluate claims against given data as a part of general multimodal fact-checking benchmarks~\cite{akhtar2023reading, akhtar2023chartcheck}.

\paragraph{Instruction-tuning across modalities and for charts}
Instruction-tuning was proposed to generalize the abilities of language models across multiple tasks~\cite{mishra-etal-2022-cross} and has become a common practice for adapting pre-trained LLMs to real-world applications\cite{Alpaca, Vicuna, instructgpt}. The success of instruction-tuning for text has led to its adoption as a standard process for multimodal VLMs too~\cite{li2023llava, zhu2023minigpt, dai2023instructblip}. Recently, domain-specific instruction-tuning has been attempted for charts that requires specially curated instruction-tuning data~\cite{han2023chartllama, masry2024chartinstruct, meng2024chartassisstant}. These methods use the underlying data tables of the chart to synthesize the instruction-tuning data. Since the data tables of charts are not capable of capturing the nuance details of charts, especially for real-world charts with complex elements, the instruction-tuning data generated using the data tables is not adequate for training models to be adept at understanding these diverse real-world charts.

\section{Conclusion and Future Work}

In the landscape of rising excitement for chart understanding and reasoning models and methods, we present ChartGemma, a multimodal model instruction-tuned on data generated directly from a diverse range of real-world chart images using a state-of-the-art backbone architecture. ChartGemma addresses two crucial shortcomings of existing instruction-tuned chart models: the instruction-tuning data is generated from the underlying data tables instead of the chart images, limiting their adaptability and extendibility to real-world, and use weakly aligned backbone models, restricting their generalizability. Our simple approach yields significant improvements over existing chart representation models, with a relatively smaller model in terms of number of parameters. Our extensive error analyses and human studies show that ChartGemma produces more realistic, informative, and factually correct outputs as compared to its contemporaries.

As future work, we aim to formulate a more diverse instruction-tuning dataset which is created using human written instructions capturing varied nuances present in charts. We also aim to propose a more generalized benchmark catered to addressing complex visual elements in charts with more chart relevant evaluation metrics.

\section*{Limitations}
Despite the effectiveness of our instruction-tuning approach and our model, there are notable limitations. Firstly, the instruction-tuning data is generated using a proprietary LLM, which could restrict the model's use in certain commercial environments. Secondly, the input resolution of our model's vision encoder is capped at 448x448; any increase in resolution leads to a quadratic rise in processing time. Third, we depend on the closed-source model, GPT4, for evaluating crucial metrics such as Informativeness and Factual Correctness. The frequent updates and potential deprecation of closed-source models pose challenges for the reproducibility of our results. Lastly, the model is prone to hallucinations, occasionally producing factually incorrect statements or erroneous code. We advise users to implement robust guardrails and exercise caution when deploying our model in real-world applications.

\section*{Ethics Statement}
Since our model generates responses autoregressively, it is prone to errors and hallucinations. The outputs can sometimes be misleading or contain inaccuracies. Additionally, there is no guarantee that the codes generated by our model will be free from malicious content. Therefore, it is crucial for users of our model to implement strict safety guidelines to mitigate these potential risks. However, the base datasets we use for further generating our instruction-tuning data are available publicly either as full datasets or URLs with public licenses. Furthermore, all chart images in our dataset were sourced from existing, publicly available research papers that have filtered out any offensive content. We plan to release our visual instruction-tuning dataset in the same way as the base datasets (images where the licenses allow us and URLs where they do not). We also release our trained ChartGemma model in easy-to-use demos and various formats and across quantizations for extremely accessible adoption by the community. For our human evaluation study, we requested the help of our research collaborators. There were no personal identification information collected during this study. As the focus of the research was about assessing models’ capabilities and limitations in several chart understanding tasks, the human evaluation performed by the authors does not add any ethical issues or unwanted biases.

\section*{Acknowledgements}

This research was supported by the Natural Sciences \& Engineering Research Council (NSERC) of Canada and Canada Foundation for Innovation (CFI). The authors acknowledge the computational resources provided by the Digital Research Alliance of Canada.

\bibliographystyle{acl_natbib}
\bibliography{main}

\begin{thebibliography}{39}
\expandafter\ifx\csname natexlab\endcsname\relax\def\natexlab#1{#1}\fi

\bibitem[{Akhtar et~al.(2023{\natexlab{a}})Akhtar, Cocarascu, and Simperl}]{akhtar2023reading}
Mubashara Akhtar, Oana Cocarascu, and Elena Simperl. 2023{\natexlab{a}}.
\newblock Reading and reasoning over chart images for evidence-based automated fact-checking.
\newblock \emph{arXiv preprint arXiv:2301.11843}.

\bibitem[{Akhtar et~al.(2023{\natexlab{b}})Akhtar, Cocarascu, and Simperl}]{akhtar-etal-2023-reading}
Mubashara Akhtar, Oana Cocarascu, and Elena Simperl. 2023{\natexlab{b}}.
\newblock \href {https://doi.org/10.18653/v1/2023.findings-eacl.30} {Reading and reasoning over chart images for evidence-based automated fact-checking}.
\newblock In \emph{Findings of the Association for Computational Linguistics: EACL 2023}, pages 399--414, Dubrovnik, Croatia. Association for Computational Linguistics.

\bibitem[{Akhtar et~al.(2023{\natexlab{c}})Akhtar, Subedi, Gupta, Tahmasebi, Cocarascu, and Simperl}]{akhtar2023chartcheck}
Mubashara Akhtar, Nikesh Subedi, Vivek Gupta, Sahar Tahmasebi, Oana Cocarascu, and Elena Simperl. 2023{\natexlab{c}}.
\newblock Chartcheck: An evidence-based fact-checking dataset over real-world chart images.
\newblock \emph{arXiv preprint arXiv:2311.07453}.

\bibitem[{{Alpaca}(2023)}]{Alpaca}
{Alpaca}. 2023.
\newblock {Alpaca}.
\newblock \url{https://crfm.stanford.edu/2023/03/13/alpaca.html}.

\bibitem[{Beyer et~al.(2024)Beyer, Steiner, Pinto, Kolesnikov, Wang, Salz, Neumann, Alabdulmohsin, Tschannen, Bugliarello, Unterthiner, Keysers, Koppula, Liu, Grycner, Gritsenko, Houlsby, Kumar, Rong, Eisenschlos, Kabra, Bauer, Bošnjak, Chen, Minderer, Voigtlaender, Bica, Balazevic, Puigcerver, Papalampidi, Henaff, Xiong, Soricut, Harmsen, and Zhai}]{beyer2024paligemma}
Lucas Beyer, Andreas Steiner, André~Susano Pinto, Alexander Kolesnikov, Xiao Wang, Daniel Salz, Maxim Neumann, Ibrahim Alabdulmohsin, Michael Tschannen, Emanuele Bugliarello, Thomas Unterthiner, Daniel Keysers, Skanda Koppula, Fangyu Liu, Adam Grycner, Alexey Gritsenko, Neil Houlsby, Manoj Kumar, Keran Rong, Julian Eisenschlos, Rishabh Kabra, Matthias Bauer, Matko Bošnjak, Xi~Chen, Matthias Minderer, Paul Voigtlaender, Ioana Bica, Ivana Balazevic, Joan Puigcerver, Pinelopi Papalampidi, Olivier Henaff, Xi~Xiong, Radu Soricut, Jeremiah Harmsen, and Xiaohua Zhai. 2024.
\newblock {PaliGemma: A versatile 3B VLM for transfer}.
\newblock \emph{arXiv preprint arXiv:2407.07726}.

\bibitem[{Callison-Burch et~al.(2006)Callison-Burch, Osborne, and Koehn}]{callison-burch-etal-2006-evaluating}
Chris Callison-Burch, Miles Osborne, and Philipp Koehn. 2006.
\newblock \href {https://aclanthology.org/E06-1032} {Re-evaluating the role of {B}leu in machine translation research}.
\newblock In \emph{11th Conference of the {E}uropean Chapter of the Association for Computational Linguistics}, pages 249--256, Trento, Italy. Association for Computational Linguistics.

\bibitem[{Chiang et~al.(2023)Chiang, Li, Lin, Sheng, Wu, Zhang, Zheng, Zhuang, Zhuang, Gonzalez, Stoica, and Xing}]{Vicuna}
Wei-Lin Chiang, Zhuohan Li, Zi~Lin, Ying Sheng, Zhanghao Wu, Hao Zhang, Lianmin Zheng, Siyuan Zhuang, Yonghao Zhuang, Joseph~E. Gonzalez, Ion Stoica, and Eric~P. Xing. 2023.
\newblock \href {https://lmsys.org/blog/2023-03-30-vicuna/} {Vicuna: An open-source chatbot impressing gpt-4 with 90\%* chatgpt quality}.

\bibitem[{Choi et~al.(2019)Choi, Jung, Park, Choo, and Elmqvist}]{Choi2019VisualizingFT}
J.~Choi, Sanghun Jung, Deok~Gun Park, J.~Choo, and N.~Elmqvist. 2019.
\newblock Visualizing for the non‐visual: Enabling the visually impaired to use visualization.
\newblock \emph{Computer Graphics Forum}, 38.

\bibitem[{Dai et~al.(2023)Dai, Li, Li, Tiong, Zhao, Wang, Li, Fung, and Hoi}]{dai2023instructblip}
Wenliang Dai, Junnan Li, Dongxu Li, Anthony Meng~Huat Tiong, Junqi Zhao, Weisheng Wang, Boyang Li, Pascale Fung, and Steven Hoi. 2023.
\newblock \href {http://arxiv.org/abs/2305.06500} {Instructblip: Towards general-purpose vision-language models with instruction tuning}.

\bibitem[{Gao et~al.(2022)Gao, Madaan, Zhou, Alon, Liu, Yang, Callan, and Neubig}]{gao2022pal}
Luyu Gao, Aman Madaan, Shuyan Zhou, Uri Alon, Pengfei Liu, Yiming Yang, Jamie Callan, and Graham Neubig. 2022.
\newblock Pal: Program-aided language models.
\newblock \emph{arXiv preprint arXiv:2211.10435}.

\bibitem[{Han et~al.(2023)Han, Zhang, Chen, Yang, Wang, Yu, Fu, and Zhang}]{han2023chartllama}
Yucheng Han, Chi Zhang, Xin Chen, Xu~Yang, Zhibin Wang, Gang Yu, Bin Fu, and Hanwang Zhang. 2023.
\newblock Chartllama: A multimodal llm for chart understanding and generation.
\newblock \emph{arXiv preprint arXiv:2311.16483}.

\bibitem[{Hoque et~al.(2022)Hoque, Kavehzadeh, and Masry}]{hoque2022chartSurvey}
Enamul Hoque, Parsa Kavehzadeh, and Ahmed Masry. 2022.
\newblock \href {https://doi.org/10.1111/cgf.14573} {Chart question answering: State of the art and future directions}.
\newblock \emph{Journal of Computer Graphics Forum (Proc. EuroVis)}, pages 555--572.

\bibitem[{Kantharaj et~al.(2022)Kantharaj, Do, Leong, Tan, Hoque, and Joty}]{open-CQA}
Shankar Kantharaj, Xuan~Long Do, Rixie Tiffany~Ko Leong, Jia~Qing Tan, Enamul Hoque, and Shafiq Joty. 2022.
\newblock Opencqa: Open-ended question answering with charts.
\newblock In \emph{Proceedings of EMNLP (to appear)}.

\bibitem[{Lee et~al.(2022)Lee, Joshi, Turc, Hu, Liu, Eisenschlos, Khandelwal, Shaw, Chang, and Toutanova}]{lee2022pix2struct}
Kenton Lee, Mandar Joshi, Iulia Turc, Hexiang Hu, Fangyu Liu, Julian Eisenschlos, Urvashi Khandelwal, Peter Shaw, Ming-Wei Chang, and Kristina Toutanova. 2022.
\newblock Pix2struct: Screenshot parsing as pretraining for visual language understanding.
\newblock \emph{arXiv preprint arXiv:2210.03347}.

\bibitem[{Li et~al.(2023)Li, Wong, Zhang, Usuyama, Liu, Yang, Naumann, Poon, and Gao}]{li2023llava}
Chunyuan Li, Cliff Wong, Sheng Zhang, Naoto Usuyama, Haotian Liu, Jianwei Yang, Tristan Naumann, Hoifung Poon, and Jianfeng Gao. 2023.
\newblock Llava-med: Training a large language-and-vision assistant for biomedicine in one day.
\newblock \emph{arXiv preprint arXiv:2306.00890}.

\bibitem[{Liu et~al.(2022)Liu, Piccinno, Krichene, Pang, Lee, Joshi, Altun, Collier, and Eisenschlos}]{liu2022matcha}
Fangyu Liu, Francesco Piccinno, Syrine Krichene, Chenxi Pang, Kenton Lee, Mandar Joshi, Yasemin Altun, Nigel Collier, and Julian~Martin Eisenschlos. 2022.
\newblock Matcha: Enhancing visual language pretraining with math reasoning and chart derendering.
\newblock \emph{arXiv preprint arXiv:2212.09662}.

\bibitem[{Liu et~al.(2023{\natexlab{a}})Liu, Wang, Yao, Chen, Song, Cho, Yacoob, and Yu}]{liu2023mmc}
Fuxiao Liu, Xiaoyang Wang, Wenlin Yao, Jianshu Chen, Kaiqiang Song, Sangwoo Cho, Yaser Yacoob, and Dong Yu. 2023{\natexlab{a}}.
\newblock Mmc: Advancing multimodal chart understanding with large-scale instruction tuning.
\newblock \emph{arXiv preprint arXiv:2311.10774}.

\bibitem[{Liu et~al.(2023{\natexlab{b}})Liu, Li, Wu, and Lee}]{liu2023visual}
Haotian Liu, Chunyuan Li, Qingyang Wu, and Yong~Jae Lee. 2023{\natexlab{b}}.
\newblock Visual instruction tuning.
\newblock \emph{arXiv preprint arXiv:2304.08485}.

\bibitem[{Masry and Hoque(2021)}]{ahmed-workshop-2021}
Ahmed Masry and Enamul Hoque. 2021.
\newblock Integrating image data extraction and table parsing methods for chart question answering.
\newblock \emph{Chart Question Answering Workshop, in conjunction with the Conference on Computer Vision and Pattern Recognition (CVPR)}, pages 1--5.

\bibitem[{Masry et~al.(2023)Masry, Kavehzadeh, Do, Hoque, and Joty}]{masry2023unichart}
Ahmed Masry, Parsa Kavehzadeh, Xuan~Long Do, Enamul Hoque, and Shafiq Joty. 2023.
\newblock {UniChart}: A universal vision-language pretrained model for chart comprehension and reasoning.
\newblock In \emph{Proceedings of the 2023 Conference on Empirical Methods in Natural Language Processing (to appear)}. Association for Computational Linguistics.

\bibitem[{Masry et~al.(2022{\natexlab{a}})Masry, Long, Tan, Joty, and Hoque}]{masry-etal-2022-chartqa}
Ahmed Masry, Do~Long, Jia~Qing Tan, Shafiq Joty, and Enamul Hoque. 2022{\natexlab{a}}.
\newblock \href {https://doi.org/10.18653/v1/2022.findings-acl.177} {{C}hart{QA}: A benchmark for question answering about charts with visual and logical reasoning}.
\newblock In \emph{Findings of the Association for Computational Linguistics: ACL 2022}, pages 2263--2279, Dublin, Ireland. Association for Computational Linguistics.

\bibitem[{Masry et~al.(2022{\natexlab{b}})Masry, Long, Tan, Joty, and Hoque}]{MSMmasry2022chartqa}
Ahmed Masry, Do~Xuan Long, Jia~Qing Tan, Shafiq Joty, and Enamul Hoque. 2022{\natexlab{b}}.
\newblock Chartqa: A benchmark for question answering about charts with visual and logical reasoning.
\newblock \emph{arXiv preprint arXiv:2203.10244}.

\bibitem[{Masry et~al.(2024)Masry, Shahmohammadi, Parvez, Hoque, and Joty}]{masry2024chartinstruct}
Ahmed Masry, Mehrad Shahmohammadi, Md~Rizwan Parvez, Enamul Hoque, and Shafiq Joty. 2024.
\newblock \href {http://arxiv.org/abs/2403.09028} {Chartinstruct: Instruction tuning for chart comprehension and reasoning}.

\bibitem[{Meng et~al.(2024)Meng, Shao, Lu, Gao, Zhang, Qiao, and Luo}]{meng2024chartassisstant}
Fanqing Meng, Wenqi Shao, Quanfeng Lu, Peng Gao, Kaipeng Zhang, Yu~Qiao, and Ping Luo. 2024.
\newblock Chartassisstant: A universal chart multimodal language model via chart-to-table pre-training and multitask instruction tuning.
\newblock \emph{arXiv preprint arXiv:2401.02384}.

\bibitem[{Methani et~al.(2020)Methani, Ganguly, Khapra, and Kumar}]{plotqa}
Nitesh Methani, Pritha Ganguly, Mitesh~M. Khapra, and Pratyush Kumar. 2020.
\newblock Plotqa: Reasoning over scientific plots.
\newblock In \emph{Proceedings of the IEEE/CVF Winter Conference on Applications of Computer Vision (WACV)}.

\bibitem[{Mishra et~al.(2022)Mishra, Khashabi, Baral, and Hajishirzi}]{mishra-etal-2022-cross}
Swaroop Mishra, Daniel Khashabi, Chitta Baral, and Hannaneh Hajishirzi. 2022.
\newblock \href {https://doi.org/10.18653/v1/2022.acl-long.244} {Cross-task generalization via natural language crowdsourcing instructions}.
\newblock In \emph{Proceedings of the 60th Annual Meeting of the Association for Computational Linguistics (Volume 1: Long Papers)}, pages 3470--3487, Dublin, Ireland. Association for Computational Linguistics.

\bibitem[{OpenAI(2023)}]{openai2023gpt4}
OpenAI. 2023.
\newblock \href {http://arxiv.org/abs/2303.08774} {{GPT-4 Technical Report}}.

\bibitem[{Ouyang et~al.(2022)Ouyang, Wu, Jiang, Almeida, Wainwright, Mishkin, Zhang, Agarwal, Slama, Ray, Schulman, Hilton, Kelton, Miller, Simens, Askell, Welinder, Christiano, Leike, and Lowe}]{instructgpt}
Long Ouyang, Jeff Wu, Xu~Jiang, Diogo Almeida, Carroll~L. Wainwright, Pamela Mishkin, Chong Zhang, Sandhini Agarwal, Katarina Slama, Alex Ray, John Schulman, Jacob Hilton, Fraser Kelton, Luke Miller, Maddie Simens, Amanda Askell, Peter Welinder, Paul Christiano, Jan Leike, and Ryan Lowe. 2022.
\newblock \href {https://doi.org/10.48550/ARXIV.2203.02155} {Training language models to follow instructions with human feedback}.

\bibitem[{Papineni et~al.(2002)Papineni, Roukos, Ward, and Zhu}]{papineni-etal-2002-bleu}
Kishore Papineni, Salim Roukos, Todd Ward, and Wei-Jing Zhu. 2002.
\newblock \href {https://doi.org/10.3115/1073083.1073135} {{B}leu: a method for automatic evaluation of machine translation}.
\newblock In \emph{Proceedings of the 40th Annual Meeting of the Association for Computational Linguistics}, pages 311--318, Philadelphia, Pennsylvania, USA. Association for Computational Linguistics.

\bibitem[{Post(2018)}]{post-2018-call}
Matt Post. 2018.
\newblock \href {https://doi.org/10.18653/v1/W18-6319} {A call for clarity in reporting {BLEU} scores}.
\newblock In \emph{Proceedings of the Third Conference on Machine Translation: Research Papers}, pages 186--191, Brussels, Belgium. Association for Computational Linguistics.

\bibitem[{Radford et~al.(2021)Radford, Kim, Hallacy, Ramesh, Goh, Agarwal, Sastry, Askell, Mishkin, Clark, Krueger, and Sutskever}]{clipvisual}
Alec Radford, Jong~Wook Kim, Chris Hallacy, Aditya Ramesh, Gabriel Goh, Sandhini Agarwal, Girish Sastry, Amanda Askell, Pamela Mishkin, Jack Clark, Gretchen Krueger, and Ilya Sutskever. 2021.
\newblock \href {http://arxiv.org/abs/2103.00020} {Learning transferable visual models from natural language supervision}.

\bibitem[{Shankar et~al.(2022)Shankar, Rixie Tiffany~Ko, Xiang, Ahmed, Megh, Enamul, and Shafiq}]{chart-to-text-acl}
Kantharaj Shankar, Leong Rixie Tiffany~Ko, Lin Xiang, Masry Ahmed, Thakkar Megh, Hoque Enamul, and Joty Shafiq. 2022.
\newblock Chart-to-text: A large-scale benchmark for chart summarization.
\newblock In \emph{In Proceedings of the Annual Meeting of the Association for Computational Linguistics (ACL), 2022}.

\bibitem[{Smith et~al.(2016)Smith, Hardmeier, and Tiedemann}]{smith-etal-2016-climbing}
Aaron Smith, Christian Hardmeier, and Joerg Tiedemann. 2016.
\newblock \href {https://aclanthology.org/W16-3414} {Climbing mont {BLEU}: The strange world of reachable high-{BLEU} translations}.
\newblock In \emph{Proceedings of the 19th Annual Conference of the {E}uropean Association for Machine Translation}, pages 269--281.

\bibitem[{Team et~al.(2023)Team, Anil, Borgeaud, Wu, Alayrac, and et~al.}]{geminiteam2023gemini}
Gemini Team, Rohan Anil, Sebastian Borgeaud, Yonghui Wu, Jean-Baptiste Alayrac, and Jiahui~Yu et~al. 2023.
\newblock \href {http://arxiv.org/abs/2312.11805} {Gemini: A family of highly capable multimodal models}.

\bibitem[{Team et~al.(2024)Team, Mesnard, Hardin, Dadashi, Bhupatiraju, Pathak, Sifre, Rivière, Kale, Love, Tafti, Hussenot, Sessa, Chowdhery, Roberts, Barua, Botev, Castro-Ros, Slone, Héliou, Tacchetti, Bulanova, Paterson, Tsai, Shahriari, Lan, Choquette-Choo, Crepy, Cer, Ippolito, Reid, Buchatskaya, Ni, Noland, Yan, Tucker, Muraru, Rozhdestvenskiy, Michalewski, Tenney, Grishchenko, Austin, Keeling, Labanowski, Lespiau, Stanway, Brennan, Chen, Ferret, Chiu, Mao-Jones, Lee, Yu, Millican, Sjoesund, Lee, Dixon, Reid, Mikuła, Wirth, Sharman, Chinaev, Thain, Bachem, Chang, Wahltinez, Bailey, Michel, Yotov, Chaabouni, Comanescu, Jana, Anil, McIlroy, Liu, Mullins, Smith, Borgeaud, Girgin, Douglas, Pandya, Shakeri, De, Klimenko, Hennigan, Feinberg, Stokowiec, hui Chen, Ahmed, Gong, Warkentin, Peran, Giang, Farabet, Vinyals, Dean, Kavukcuoglu, Hassabis, Ghahramani, Eck, Barral, Pereira, Collins, Joulin, Fiedel, Senter, Andreev, and Kenealy}]{gemma}
Gemma Team, Thomas Mesnard, Cassidy Hardin, Robert Dadashi, Surya Bhupatiraju, Shreya Pathak, Laurent Sifre, Morgane Rivière, Mihir~Sanjay Kale, Juliette Love, Pouya Tafti, Léonard Hussenot, Pier~Giuseppe Sessa, Aakanksha Chowdhery, Adam Roberts, Aditya Barua, Alex Botev, Alex Castro-Ros, Ambrose Slone, Amélie Héliou, Andrea Tacchetti, Anna Bulanova, Antonia Paterson, Beth Tsai, Bobak Shahriari, Charline~Le Lan, Christopher~A. Choquette-Choo, Clément Crepy, Daniel Cer, Daphne Ippolito, David Reid, Elena Buchatskaya, Eric Ni, Eric Noland, Geng Yan, George Tucker, George-Christian Muraru, Grigory Rozhdestvenskiy, Henryk Michalewski, Ian Tenney, Ivan Grishchenko, Jacob Austin, James Keeling, Jane Labanowski, Jean-Baptiste Lespiau, Jeff Stanway, Jenny Brennan, Jeremy Chen, Johan Ferret, Justin Chiu, Justin Mao-Jones, Katherine Lee, Kathy Yu, Katie Millican, Lars~Lowe Sjoesund, Lisa Lee, Lucas Dixon, Machel Reid, Maciej Mikuła, Mateo Wirth, Michael Sharman, Nikolai Chinaev, Nithum Thain, Olivier Bachem,
  Oscar Chang, Oscar Wahltinez, Paige Bailey, Paul Michel, Petko Yotov, Rahma Chaabouni, Ramona Comanescu, Reena Jana, Rohan Anil, Ross McIlroy, Ruibo Liu, Ryan Mullins, Samuel~L Smith, Sebastian Borgeaud, Sertan Girgin, Sholto Douglas, Shree Pandya, Siamak Shakeri, Soham De, Ted Klimenko, Tom Hennigan, Vlad Feinberg, Wojciech Stokowiec, Yu~hui Chen, Zafarali Ahmed, Zhitao Gong, Tris Warkentin, Ludovic Peran, Minh Giang, Clément Farabet, Oriol Vinyals, Jeff Dean, Koray Kavukcuoglu, Demis Hassabis, Zoubin Ghahramani, Douglas Eck, Joelle Barral, Fernando Pereira, Eli Collins, Armand Joulin, Noah Fiedel, Evan Senter, Alek Andreev, and Kathleen Kenealy. 2024.
\newblock \href {http://arxiv.org/abs/2403.08295} {Gemma: Open models based on gemini research and technology}.

\bibitem[{Vaswani et~al.(2017)Vaswani, Shazeer, Parmar, Uszkoreit, Jones, Gomez, Kaiser, and Polosukhin}]{vaswani2017attention}
Ashish Vaswani, Noam Shazeer, Niki Parmar, Jakob Uszkoreit, Llion Jones, Aidan~N Gomez, {\L}ukasz Kaiser, and Illia Polosukhin. 2017.
\newblock Attention is all you need.
\newblock In \emph{Advances in neural information processing systems}, pages 5998--6008.

\bibitem[{Zhai et~al.(2023)Zhai, Mustafa, Kolesnikov, and Beyer}]{siglip}
Xiaohua Zhai, Basil Mustafa, Alexander Kolesnikov, and Lucas Beyer. 2023.
\newblock \href {http://arxiv.org/abs/2303.15343} {Sigmoid loss for language image pre-training}.

\bibitem[{Zheng et~al.(2023)Zheng, Chiang, Sheng, Zhuang, Wu, Zhuang, Lin, Li, Li, Xing, Zhang, Gonzalez, and Stoica}]{zheng2023judging}
Lianmin Zheng, Wei-Lin Chiang, Ying Sheng, Siyuan Zhuang, Zhanghao Wu, Yonghao Zhuang, Zi~Lin, Zhuohan Li, Dacheng Li, Eric Xing, Hao Zhang, Joseph~E. Gonzalez, and Ion Stoica. 2023.
\newblock \href {https://openreview.net/forum?id=uccHPGDlao} {Judging {LLM}-as-a-judge with {MT}-bench and chatbot arena}.
\newblock In \emph{Thirty-seventh Conference on Neural Information Processing Systems Datasets and Benchmarks Track}.

\bibitem[{Zhu et~al.(2023)Zhu, Chen, Shen, Li, and Elhoseiny}]{zhu2023minigpt}
Deyao Zhu, Jun Chen, Xiaoqian Shen, Xiang Li, and Mohamed Elhoseiny. 2023.
\newblock Minigpt-4: Enhancing vision-language understanding with advanced large language models.
\newblock \emph{arXiv preprint arXiv:2304.10592}.

\end{thebibliography}
\newpage

\appendix

\begin{appendices}

\section{Chart Instruction Data Generation}
\label{appendix:graph}
\subsection{Chart Corpora Collection}
\label{appendix:webchart}

We collect chart across 3 categories based on their source and method of generation as mentioned in \cref{subsec:chart_corpora_collection}. We show the exact statistics and sources under each category in \cref{tab:instructiondata}.

\begin{table*}[t]
 \setlength\extrarowheight{1pt}
 \centering
 \setlength{\tabcolsep}{10pt}

 \scalebox{0.5}{
 \begin{tabular}{lccccc|ccccc|c}
  
  \toprule

   & \multicolumn{5}{c}{Predefined Tasks} & \multicolumn{4}{c}{Open Ended Tasks} \\
  
  Dataset
   & \makecell{CoT \\ Reasoning} & \makecell{Chart \\ Summarization} & 
   \makecell{Fact \\ Checking} & \makecell{Chart-to \\Markdown}  & \makecell{Coding \\ Abilities}& \makecell{Trend \\ Analysis} & \makecell{Data \\ Comparison} & \makecell{Data \\ Interpretation} & \makecell{Data \\ Visualization} & Others & \#Charts \\%{Model generated} \\

  \midrule

  \multicolumn{8}{l}{\textbf{Synthetic Sources}} \\
  PlotQA   & - & - & - 
  & 5000  & - & - & - & - & - & - & 5000 \\

  ChartFC   & - & -   
  & 28000 & -  & - & - & - & - & - & - & 12702\\

  \midrule
  \multicolumn{8}{l}{\textbf{Specialized Websites}} \\
  Statista   & 2688 & 4996 & 1296 
  & 2377   & 42098 & 334 & 172 & 373 & 231 & 3027 & 19748\\ 

  Pew   & 11951 & 4999 & 1251 
  & 1784  & 10034 & 281 & 290 & 307 & 129 & 2873 & 7401\\

  OECD   & 243 & 500   
  & 644 & 20838  & 357 & 39 & 47 & 69 & 31 & 489 & 21712\\

  OWID   & 717 & 500   
  & 375 & 2285  & 1490 & 40 & 38 & 61 & 28 & 547 & 3803 \\

  ChartCheck (Wikipedia)   & - & 1527   
  & 7603 & -  & - & 98 & 96 & 178 & 65 & 1642 & 1530 \\

  \midrule
  \multicolumn{8}{l}{\textbf{General Web}} \\
  
  WebCharts   & 10576 & 50046 & 6434 
  & 18216  & 3400 & 4331 & 6283 & 4680 & 1785 & 51436 & 50961 \\

  \midrule
  Total  & 26,175 & 62,241 &  45,603 & 22,603 & 57,379 & 792 & 6926 & 988 & 2269 & 60,014 & 122,857
  \\ \bottomrule

 \end{tabular}
 }
 \caption{\small {The number of generated examples for each %
 tasks based on data samples of the mentioned dataset. Some of the charts are used in multiple tasks. In the last column, we show the number of distinct charts used for instruction generation samples.
 }
 }
 \label{tab:instructiondata}
\end{table*}

\paragraph{Sources for instruction-tuning tasks}
For the pre-defined tasks used for generating instruction-tuning data, we also augment the instructions generated by the multimodal LLM with the training sets of existing benchmark datasets. 

\subsection{Instruction Dataset Analysis}
\label{appendix:data_analysis}

\begin{figure*}[t!]
    \centering
    \scalebox{0.98}{\includegraphics[width=1\textwidth]{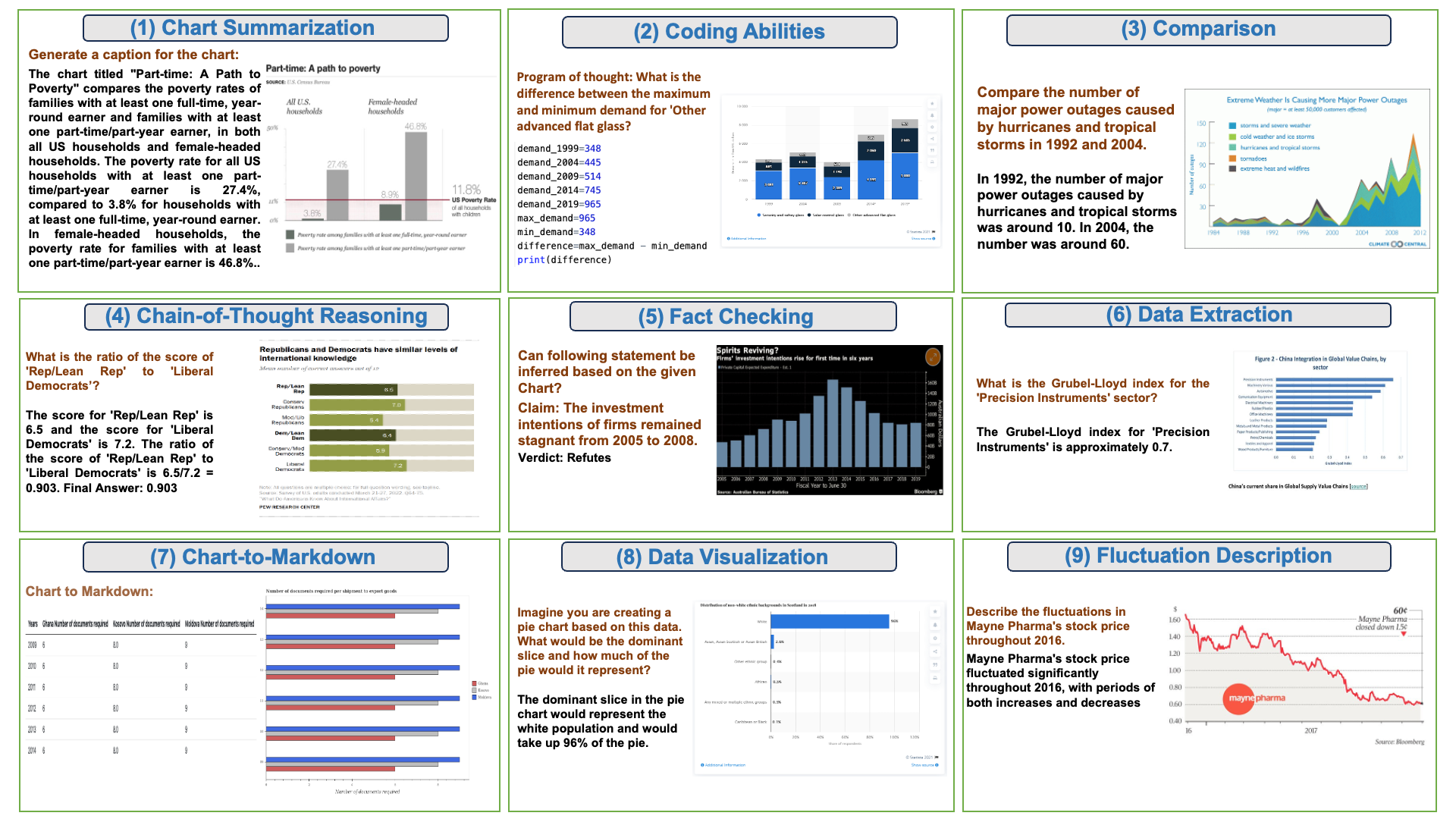}}
    \caption{Diverse examples from our visual instruction-tuning tasks that focuses on the visual attributes of the chart images which are highlighted in green.  
    }
    \label{fig:visual_examples2}
\end{figure*}

\begin{figure*}[t!]
    \centering
    \scalebox{0.98}{\includegraphics[width=1\textwidth]{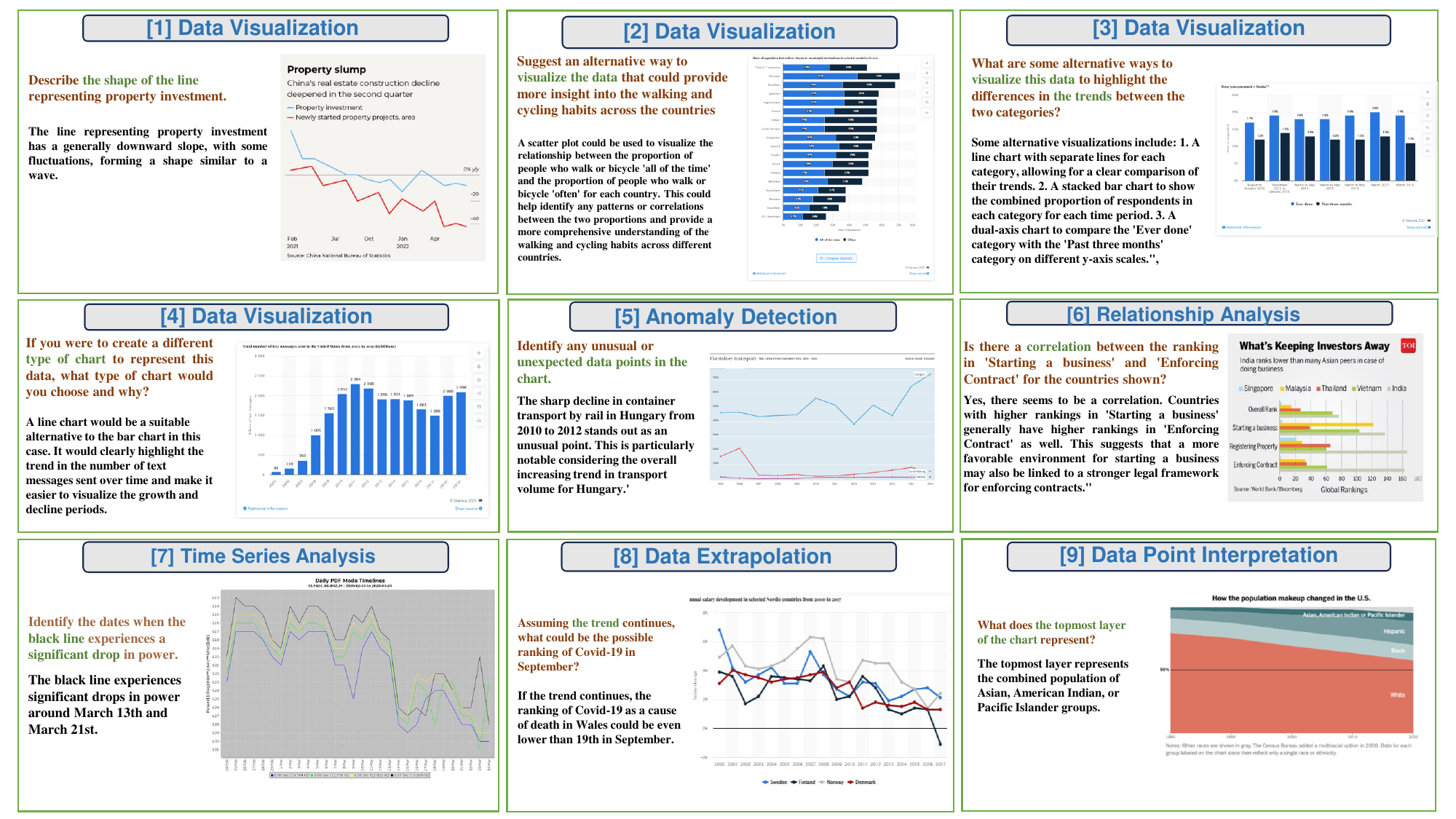}}
    \caption{Diverse examples from our open-ended instruction-tuning tasks that focuses on the visual attributes of the chart images which are highlighted in green.  
    }
    \label{fig:visual_examples}
\end{figure*}

Our instruction-tuning dataset comprises of both closed-ended response generation and open-ended answering. \cref{fig:visual_examples2} shows diverse visual instruction-tuning tasks that are generally inspired from existing chart evaluation benchmarks, and \cref{fig:visual_examples} shows diverse visual instruction-tuning tasks inspired from open-ended chart understanding and reasoning.

\paragraph{Instruction-tuning dataset quality} As mentioned in \cref{subsec:dataset_characteristics}, our instruction-tuning dataset's instructions accurately reflect the chart content approximately 82\% of the times, and are partially correct 8\% times. We present some examples where our instructions are correct and incorrect in \cref{tab:appendix-comparison-correct} and partially correct in \cref{tab:appendix-comparison-partially-correct}.

\begin{table*}[t]
\centering
\scalebox{0.50}{
\begin{tabular}{p{0.5\linewidth}p{0.3\linewidth}p{0.4\linewidth}p{0.2\linewidth}}
\Xhline{4\arrayrulewidth}
\textbf{Chart Image} & \textbf{Chart Instructions} & \textbf{Answers}  & \textbf{Label}\\ \hline \hline
\multicolumn{1}{p{0.5\linewidth}|}{\raisebox{-\height}{\includegraphics[width=0.5\textwidth]{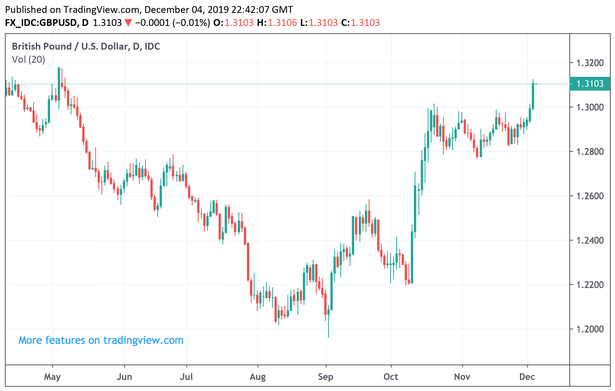}}} &
Based on the chart, what can you conclude about the performance of the British Pound against the US Dollar in the last six months?
 & The British Pound has depreciated against the US Dollar over the past six months, with a significant drop between May and August and a slight recovery towards the end of the period.
 & Correct
 \\ \midrule
\multicolumn{1}{p{0.5\linewidth}|}{\raisebox{-\height}{\includegraphics[width=0.5\textwidth]{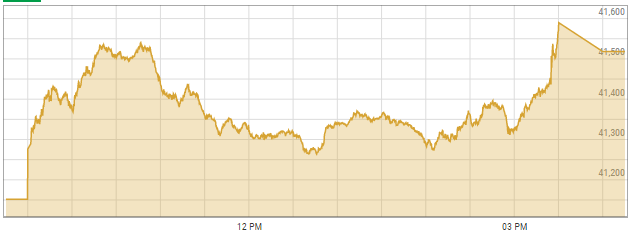}}} &
Based on the trend observed in the chart, what would you predict the value of the data to be at 4:00 PM?

& Based on the trend, the value of the data at 4:00 PM would likely be slightly lower than the value at 3:00 PM, possibly around 41,400.
 & Correct
\\ \midrule
\multicolumn{1}{p{0.5\linewidth}|}{\raisebox{-\height}{\includegraphics[width=0.5\textwidth]{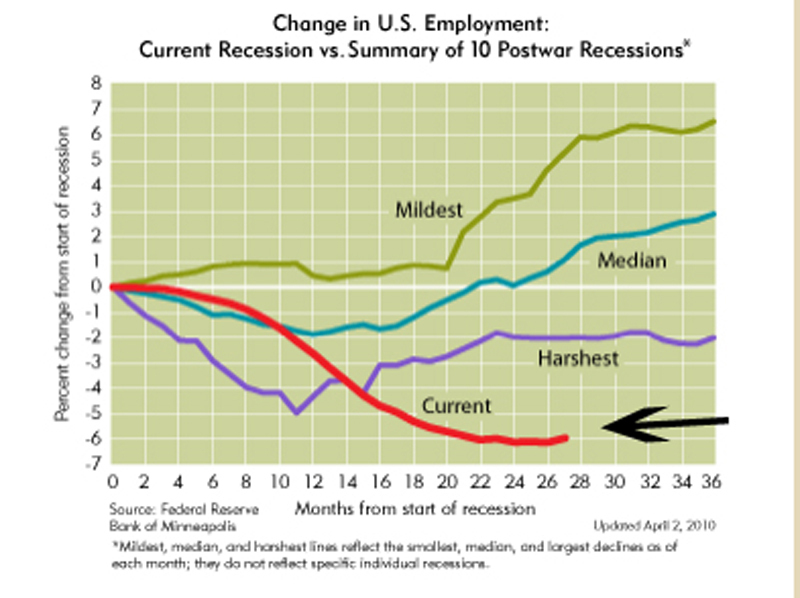}}} &
Based on the trend of the 'Current' recession line, what is the predicted percent change from the start of the recession after 35 months?

& It is difficult to predict the exact percent change after 35 months based solely on the trend of the 'Current' recession line. However, based on the upward trend after 28 months, it is likely to be around -5\%.
 & Correct
\\ \midrule
\multicolumn{1}{p{0.5\linewidth}|}{\raisebox{-\height}{\includegraphics[width=0.5\textwidth]{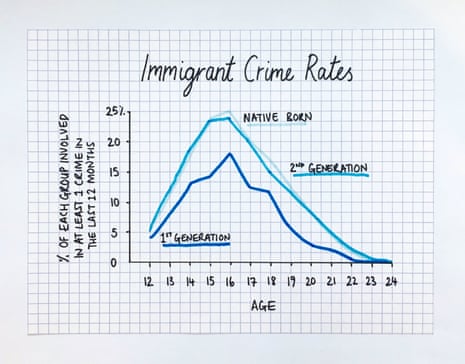}}} &
Based on the chart, what might be a possible explanation for the high crime rate among native-born individuals?

& It is difficult to say for sure based on this chart alone. However, some potential explanations could be socioeconomic factors, access to resources, or cultural influences.
 & Correct

 \\ \midrule
\multicolumn{1}{p{0.5\linewidth}|}{\raisebox{-\height}{\includegraphics[width=0.5\textwidth]{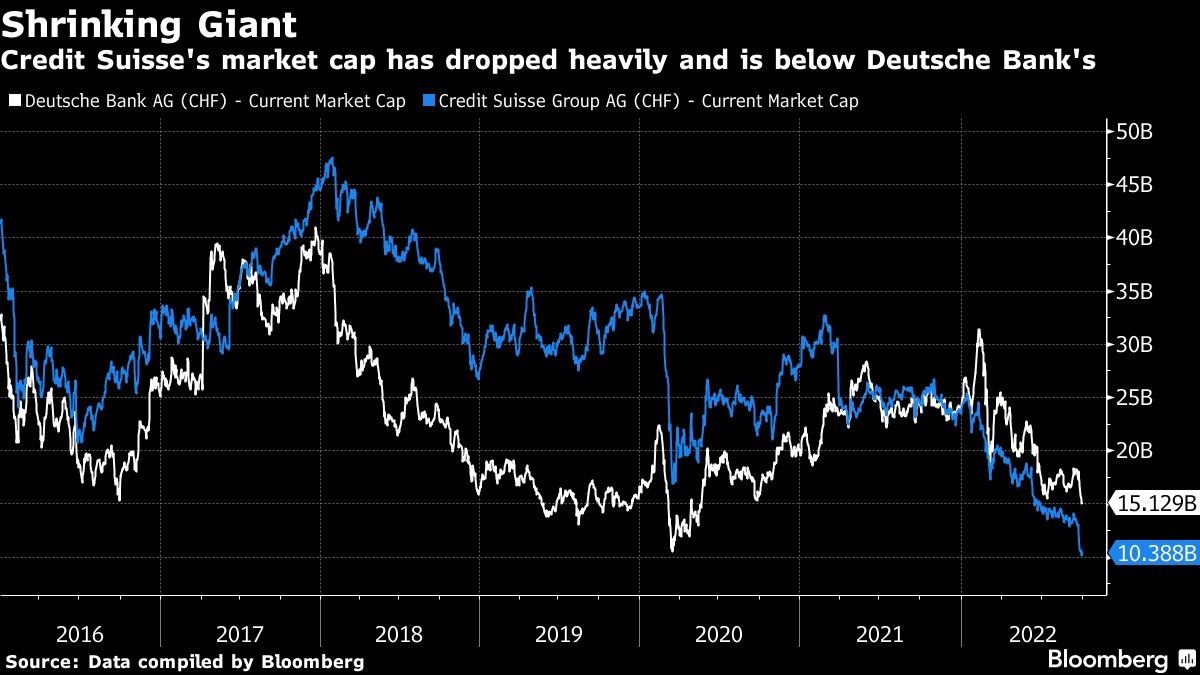}}} &
Compare the market cap of Deutsche Bank and Credit Suisse in 2021.

& Deutsche Bank and Credit Suisse in 2021.	\textcolor{red}{In 2021, Deutsche Bank's market cap was significantly higher than Credit Suisse's. Deutsche Bank's market cap was around 25 billion while Credit Suisse's was around 15 billion.}
 & Incorrect

 \\ \midrule
\multicolumn{1}{p{0.5\linewidth}|}{\raisebox{-\height}{\includegraphics[width=0.5\textwidth]{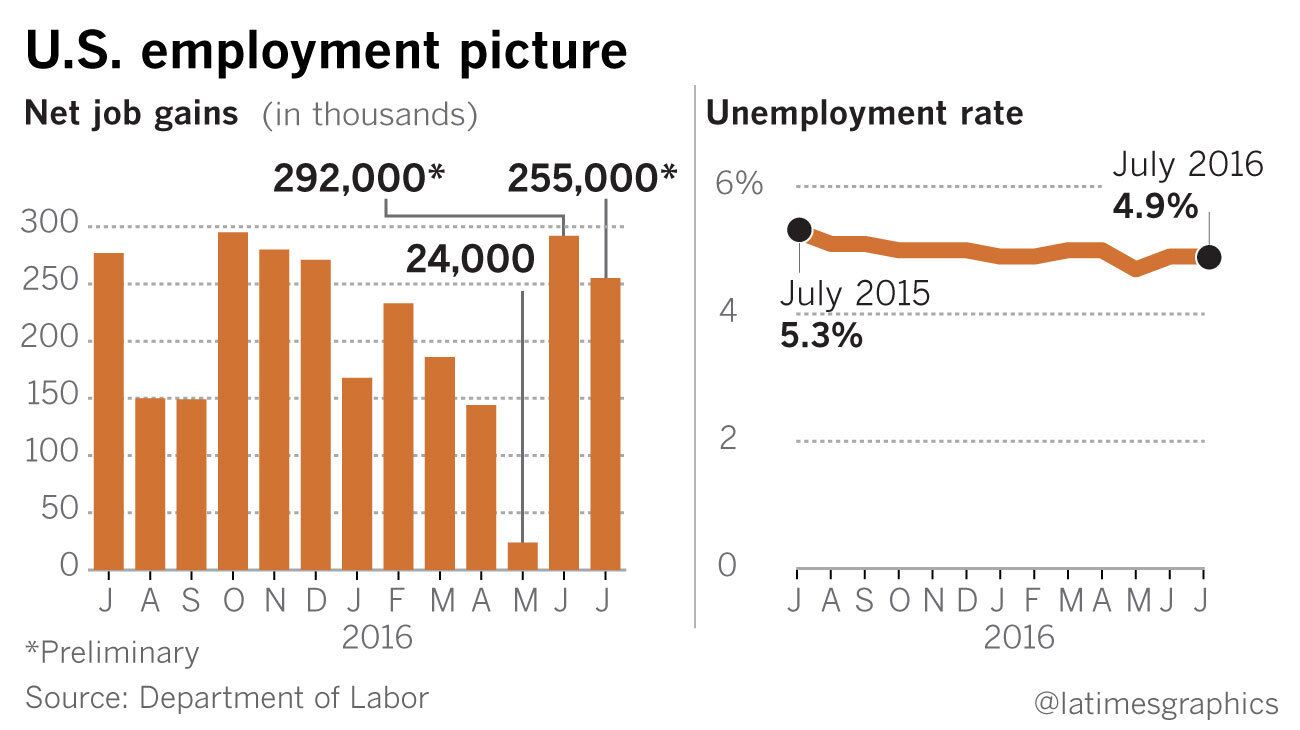}}} &
Describe the trend of job gains in the chart.

& The trend of job gains is generally increasing, with a few dips in between. It experienced a \textcolor{red}{significant peak in July 2016 and May 2016.}
 & Incorrect

  \\ \midrule
\multicolumn{1}{p{0.5\linewidth}|}{\raisebox{-\height}{\includegraphics[width=0.5\textwidth]{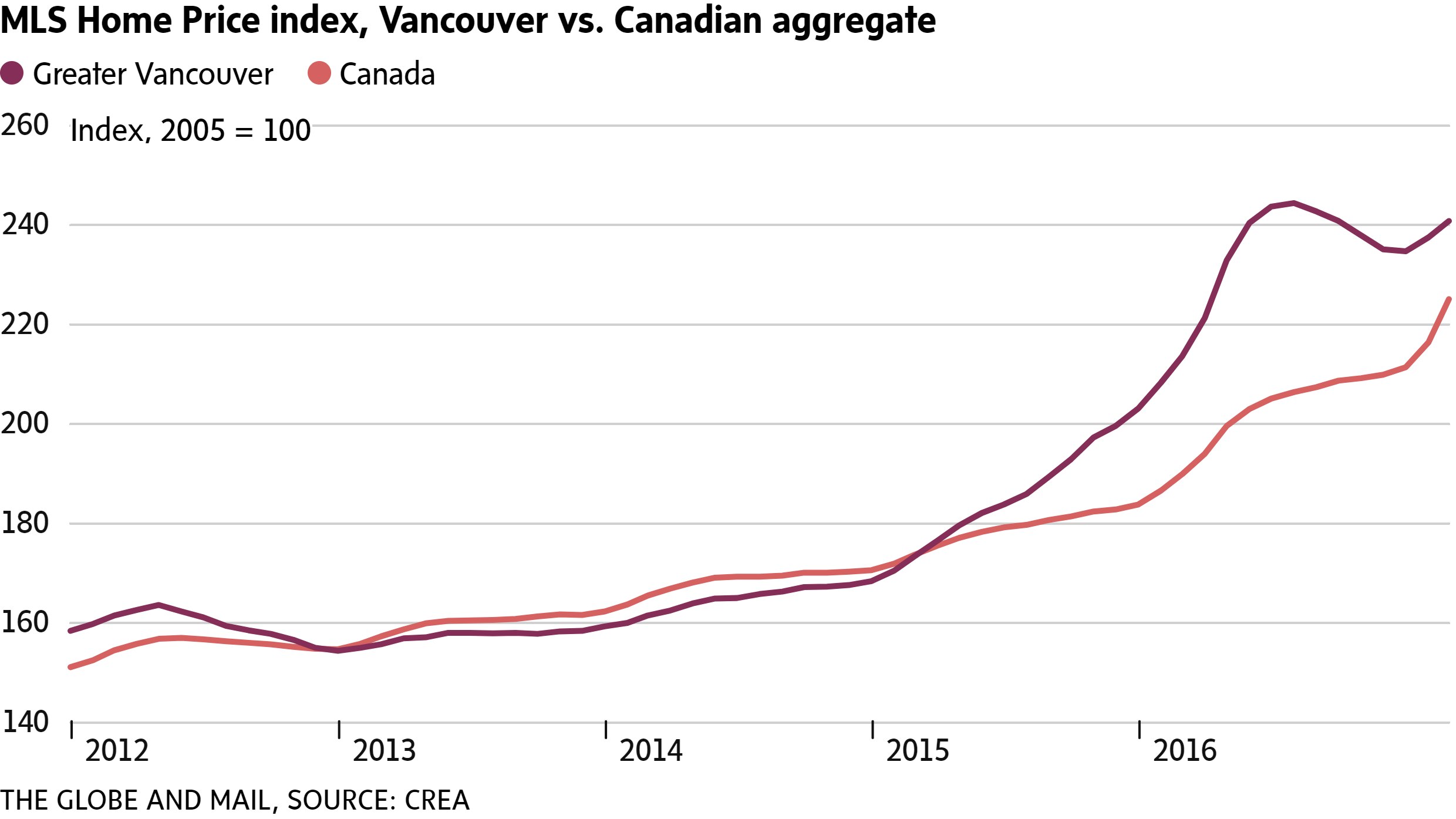}}} &
Compare the home price index in Greater Vancouver and Canada in 2015.

& In 2015, the home price index in Greater Vancouver was significantly higher than in Canada. \textcolor{red}{While Greater Vancouver's index was around 180, Canada's index was around 170.}
 & Incorrect
\\ \hline
\end{tabular}
}
\caption{\label{tab:appendix-comparison-correct}Example answers generated from Gemini Flash 1.5 based on the instructions supplied. We present some correct generations and some incorrect generations (with highlights in \textcolor{red}{red}).
}
\vskip -1ex
\end{table*}

\begin{table*}[t]
\centering
\scalebox{0.50}{
\begin{tabular}{p{0.5\linewidth}p{0.3\linewidth}p{0.4\linewidth}p{0.2\linewidth}}
\Xhline{4\arrayrulewidth}
\textbf{Chart Image} & \textbf{Chart Instructions} & \textbf{Answers}  & \textbf{Label}\\ \hline \hline

\multicolumn{1}{p{0.5\linewidth}|}{\raisebox{-\height}{\includegraphics[width=0.5\textwidth]{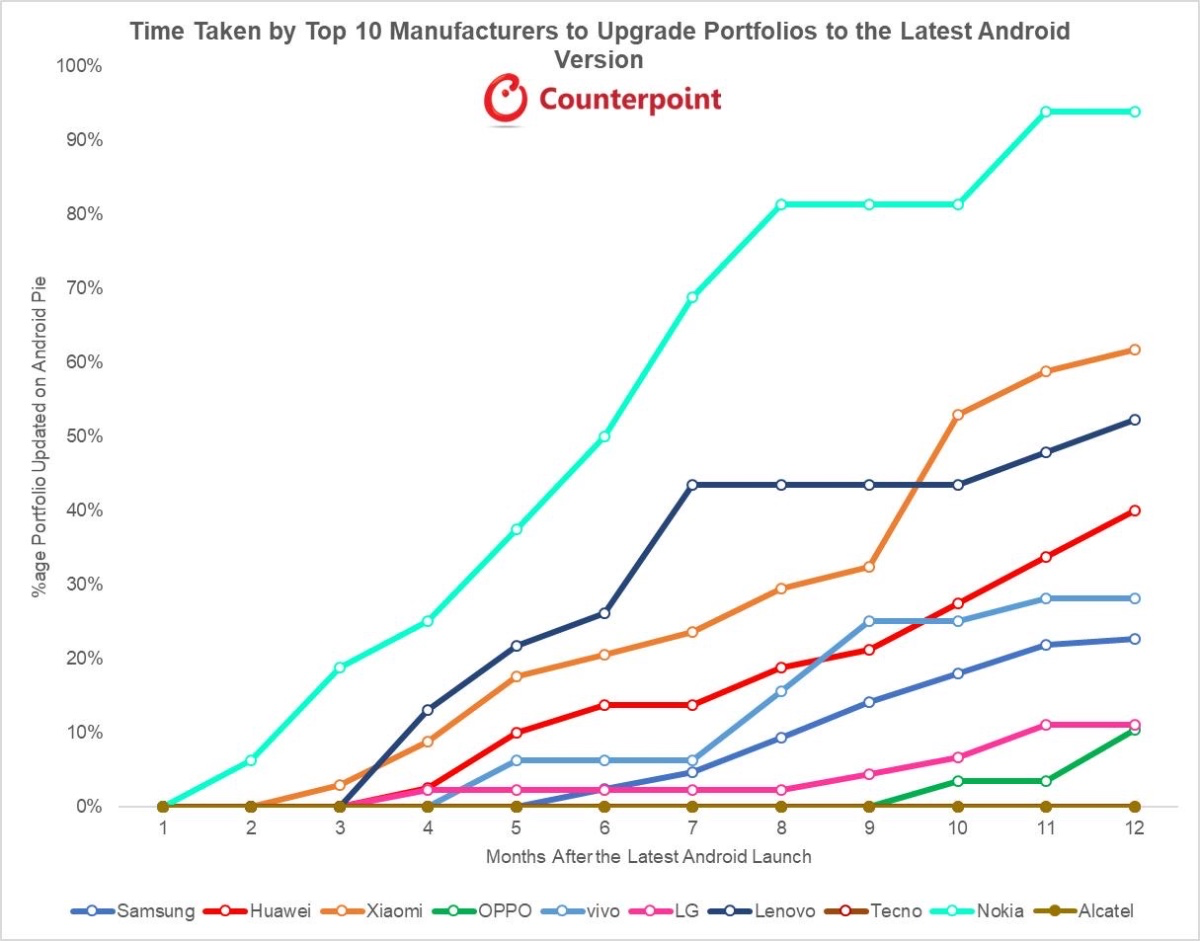}}} &
What percentage of Samsung's portfolio was updated to the latest Android version after 9 months?

& \textcolor{amber}{Around 25\% of Samsung's portfolio was updated to the latest Android version after 9 months.}
 & Partially correct

 \\ \midrule
\multicolumn{1}{p{0.5\linewidth}|}{\raisebox{-\height}{\includegraphics[width=0.5\textwidth]{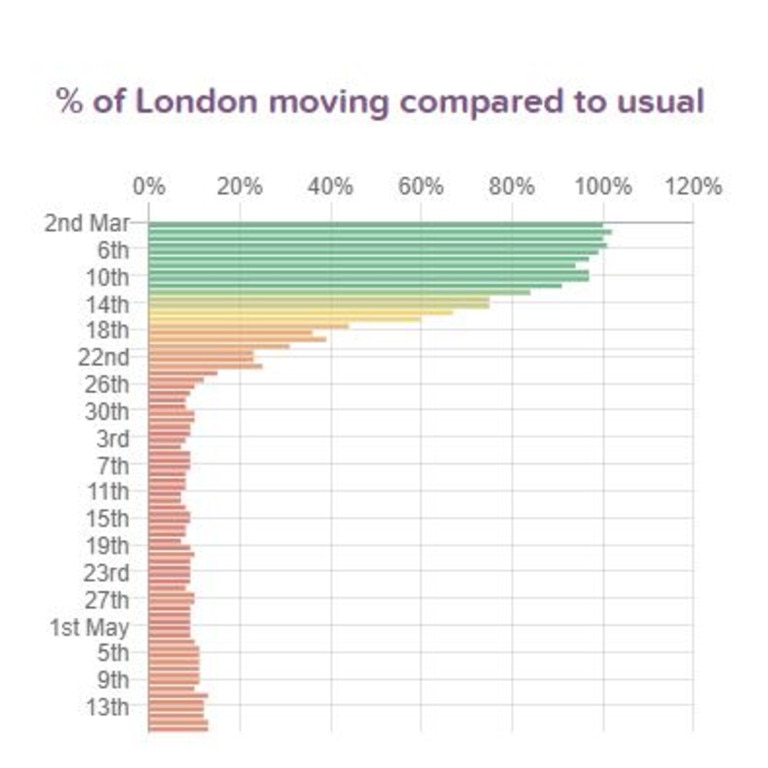}}} &
Identify the dates when the percentage of Londoners moving compared to usual was below 30%

& The dates when the percentage was below 30\% are: March 3rd, March 7th, March 11th, March 15th, March 19th, March 23rd, March 27th, 1st May, 5th May, 9th May, and 13th May. \textcolor{amber}{Missing dates}
 & Partially correct

\\ \hline
\end{tabular}
}
\caption{\label{tab:appendix-comparison-partially-correct}Example answers generated from Gemini Flash 1.5 based on the instructions supplied. We present some partially correct generations here (with highlights in \textcolor{amber}{orange}).
}
\vskip -1ex
\end{table*}

\newtcolorbox[auto counter, number within=section]{AgentPrompt}[2][]{%
    enhanced,
    colframe=gray,
    colback=white,
    boxrule=0.5mm,
    width=\textwidth,
    attach boxed title to top left={yshift=-2mm,xshift=2mm},
    boxed title style={colframe=gray, colback=white, rounded corners},
    title=#2,
    fonttitle=\bfseries\color{black},
    #1
}

\subsection{Prompt Templates for Instruction-tuning Data Generation}

We present the prompt templates provided to Gemini Flash-1.5 to generate instruction-tuning data for the program-aided design task in \cref{fig:pad-prompt} and an open-ended task in \cref{fig:open-ended-prompt}. Our prompt templates draw inspiration from the templates used in ChartInstruct~\cite{masry2024chartinstruct} and the ChartQA prompt used in Gemini Flash~\cite{geminiteam2023gemini}.

\begin{figure*}
    \centering
    \begin{AgentPrompt}{Example Prompt - Generate Instruction-tuning data for Program-Aided Design}

\begin{tiny}
\begin{verbatim}
Generate numerical and visual question-answer pairs for an LLM that we are trying to tune for Chart Numerical and Visual 
Reasoning. Your response should be in a json format where each example has three fields: input: which only asks a 
numerical/visual question, program of thought: a python program that can be executed to produce the final answer, and 
final answer: which is the final answer to the input question based on the chart image.
For the final answer X, follow the following instructions:
* X should contain as few words as possible.
* Don’t paraphrase or reformat the text you see in the image.
* If the final answer has two or more items, provide it in the list format like [1, 2].
* When asked to give a ratio, give out the decimal value like 0.25 instead of 1:4.
* When asked to give a percentage, give out the whole value like 17 instead of decimal
like 0.17%
* Don’t include any units in the answer.
* Try to include the full label from the graph when asked about an entity.
Generate ten questions that contain some numerical operations such as, but not limited to, max, min, sum, average, 
difference, ratio, median, mode, ..etc. Generate another five questions that not only have numerical operations, but also 
some visual aspects such as leftmost, rightmost, top, bottom, middle, peak, colors, ..etc. Generate five simple data 
retrieval questions that ask about values, x-labels, or legend labels from the chart. Generate another five yes/no 
numerical reasoning questions whose answers must be either Yes or No. Generate another four questions that ask to count 
some elements in the chart (e.g., the number of bars/pie slices/colors/x-labels).
Remember that the program of thought must be an executable python code that solves the question step by step and prints 
the answer in the end.
\end{verbatim}
\end{tiny}
    \end{AgentPrompt}
    \caption{Prompt to generate instruction-tuning data for the program-aided design task using Gemini Flash-1.5.}
    \label{fig:pad-prompt}
\end{figure*}

\begin{figure*}
    \centering
    \begin{AgentPrompt}{Example Prompt - Generate Instruction-tuning data for Open-ended Tasks}

\begin{tiny}
\begin{verbatim}
Generate different instruction-tuning tasks for an LLM that we are trying to tune for Chart Understanding. Your response 
should be in a json format where each example has three fields: task type, input: which only asks a question or an 
instruction related to the task type and the given chart, and expected output: which is the answer to the input 
question/instruction based on the input information. Use the following chart image to generate 10 unique tasks
\end{verbatim}
\end{tiny}
    \end{AgentPrompt}
    \caption{Prompt to generate instruction-tuning data for open-ended tasks using Gemini Flash-1.5.}
    \label{fig:open-ended-prompt}
\end{figure*}

\section{Experiments and Results}

\subsection{Hyperparameter settings}
\label{subsec:appendix-hyperparameters}

\begin{table}[t]
 \setlength\extrarowheight{2pt}
 \centering
 \caption{
  Hyperparameters and training details of our experiments.
 }
 \scalebox{0.45}{\begin{tabular}{l|cccc}
  
    \toprule
    \midrule
   \textbf{Experiment} & \textbf{\# Epochs} & \textbf{Learning Rate} & \textbf{Batch Size} & \textbf{Hours}\\ \midrule

    & \multicolumn{4}{c}{\textbf{Instruction-tuning}}  \\ \midrule
   ChartGemma & 5 & 5e-5 & 32 & ~58 \\ \midrule

   & \multicolumn{4}{c}{\textbf{Ablations}}  \\ \midrule
   PaliGemma (chartinstruct) & 1 & 5e-5 & 32 & ~22 \\
   LLaVA + our dataset & 1 & 2e-5 & 32 & ~11 \\
   ChartGemma & 1  & 5e-5 & 32 & ~11  \\ \midrule
   
    & \multicolumn{4}{c}{\textbf{Finetuning on benchmarks}}  \\\midrule
   PaliGemma (ChartFC) & 10 & 5e-5 & 32 & ~2 \\ 
   PaliGemma (ChartCheck) & 10 & 5e-5 & 32 & ~4 \\ 
   ChartInstruct-LLama2 (ChartCheck) & 10 & 2e-5 & 32 & ~2 \\
   ChartInstruct-Flan-T5-XL (ChartCheck) & 10 & 2e-5 & 32 & ~1 \\ \midrule
   
    \bottomrule

 \end{tabular}}
 \label{tab:hyperparameters}
\end{table}

We present the hyperparameter settings for instruction-tuning and fine-tuning on the benchmarks in \cref{tab:hyperparameters}.

\subsection{Prompt templates for evaluation}

We show the prompt given to GPT4 for evaluating the outputs of the open-ended tasks, Chart2Text and our curated 'Web' set for summarization and OpenCQA in \cref{fig:gpt4_summary_eval_prompt} and \cref{fig:gpt4_opencqa_eval_prompt}, respectively.

\begin{figure*}
    \centering
    \begin{AgentPrompt}{Example Prompt - Evaluating generated summaries}

\begin{tiny}
\begin{verbatim}
You will be provided with two summaries generated by different models for chart summarization. 
Your task is to evaluate each summary based on three key factors:

Informativeness: How much useful and relevant information from the chart does the summary cover? Does it effectively 
convey the main trends and insights?
Factual Correctness: How accurate is the summary in reflecting the information presented in the chart?
Structure: How well-structured is the summary? Does it include an introduction, a body with key insights, and a 
conclusion?
You are required to assign a score from 1 to 5 for each factor, for each summary. Please provide your ratings in the 
following JSON format:
{
    'summary 1': {
        'Informativeness' : score,
        'Factual Correctness' : score,
        'Structure' : score,
    },
    'summary 2': {
        'Informativeness' : score,
        'Factual Correctness' : score,
        'Structure' : score,
    },    
}
Do not return anything else other than the json above.
\end{verbatim}
\end{tiny}
    \end{AgentPrompt}
    \caption{Example prompt to evaluate open-ended summary generation for Chart2Text and the 'Web' set of charts using GPT4.}
    \label{fig:gpt4_summary_eval_prompt}
\end{figure*}

\begin{figure*}
    \centering
    \begin{AgentPrompt}{Example Prompt - Evaluating OpenCQA}

\begin{tiny}
\begin{verbatim}
You will be provided with two answers generated by different models for a question about a chart image. 
Your task is to evaluate each answer based on three key factors:
Informativeness: How much useful and relevant information from the chart does the answer cover?
Factual Correctness: How accurate is the answer in reflecting the information presented in the chart?
Relevance: How relevant is the answer to the given question? 

You are required to assign a score from 1 to 5 for each factor, for each answer. Please provide your ratings in the 
following JSON format:
{
    'summary 1': {
        'Informativeness' : score,
        'Factual Correctness' : score,
        'Relevance' : score,
    },
    'summary 2': {
        'Informativeness' : score,
        'Factual Correctness' : score,
        'Relevance' : score,
    },    
}}
\end{verbatim}
\end{tiny}
    \end{AgentPrompt}
    \caption{Example prompt to evaluate open-ended answer generation for OpenCQA using GPT4.}
    \label{fig:gpt4_opencqa_eval_prompt}
\end{figure*}

\subsection{GPT4 evaluation on open-ended generation tasks}
\label{sec:appendix-gpt4-eval}

We show the informativeness, factual correctness, and relevance results on the open-ended generation tasks, namely 
Chart2Text(Statista and Pew), OpenCQA, and our curated 'Web' set of charts in Table \ref{tab:evaluation-gpt4}.

\begin{table}[t!]
\vspace{-0.3em}
\setlength{\tabcolsep}{3pt}
    \centering
    \small
    \scalebox{0.60}{\begin{tabular}{lccc}
    \toprule 
    & \textbf{Informativeness}  & \textbf{Factual Correctness} & \textbf{Structure} \\ \midrule
    \textbf{Statista} \\ 
    ChartInstruct-LLama2 &    3.33 &    2.96 & 3.58  \\
    
    \model &    3.65 &    3.60 &   3.66
    \\ \midrule
    \textbf{Pew} \\ 
    ChartInstruct-LLama2 &    3.38 &    3.09 & 3.65  \\
    
    \model &    4.09 &    4.36 &   3.85
    \\ \midrule
    \textbf{OpenCQA} \\ 
    ChartInstruct-LLama2 &    3.54 &    3.46 & 4.56  \\
    
    \model &    3.26 & 3.48 & 4.19
    \\ \midrule
    \textbf{Web} \\ 
    ChartInstruct-LLama2 &    3.22 &    2.68 & 3.33  \\
    
    \model &    3.29 &    3.28 &   3.76
    \\
    \bottomrule
    \end{tabular}
    }
    \vspace{-2mm}
    \captionsetup{width=0.97\linewidth}
    \caption{\small GPT4 scores (from 1-5, with 5 being the highest) on the informativeness and factual correctness of outputs generated by ChartInstruct-LLaMA2 and ChartGemma (refer to \cref{subsec:open-ended-eval}).
    }
    \label{tab:evaluation-gpt4}
\end{table}

\begin{figure}[t!]
     \centering
        \includegraphics[width=0.98\textwidth]{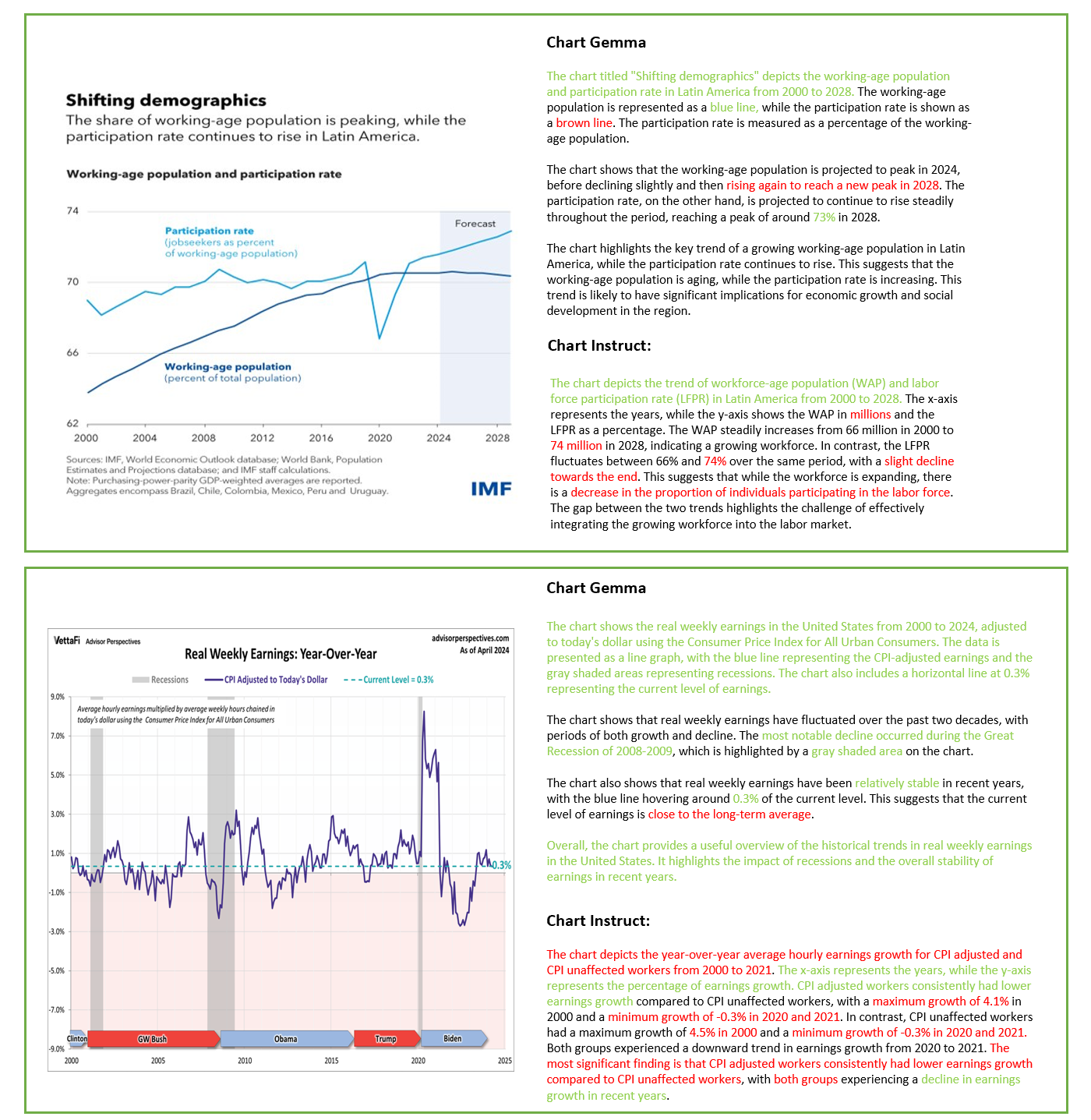}
        \vspace{-1mm}
         \caption{\small{Comparison between \model\ and ChartInstruct-LLama2 for chart captioning. } 
         }\label{fig:modelcomp}
\end{figure}

\subsection{Human Evaluation Study}
During the human evaluation study, we provided the human annotators with the same instructions used to prompt GPT4 as depicted in \cref{fig:gpt4_summary_eval_prompt} and \cref{fig:gpt4_opencqa_eval_prompt}.
We recruited two human volunteers for the study from our research lab, both were of South-east Asian (Indian subcontinent) origin and adept in the English language.

\begin{table}[t!]
\vspace{-0.3em}
\setlength{\tabcolsep}{3pt}
    \centering
    \small
    \scalebox{0.75}{\begin{tabular}{lccc}
    \toprule 
    & \textbf{Informativeness} & \textbf{Factual Correctness} & \textbf{Structure} \\ \midrule
    ChartInstruct-LLaMA2 &    3.18 &    2.80 & 3.80  \\
    ChartGemma &    3.79 &    3.59 &   3.82
    \\
    \midrule
    $p-value$ & $6.31 \times 10^{-6}$ & $2.68 \times 10^{-7}$ & $0.457$  \\
    \bottomrule
    \end{tabular}
    }
    \vspace{-2mm}
    \captionsetup{width=0.97\linewidth}
    \caption{\small Human evaluation scores on the informativeness, factual correctness, and structure of outputs generated by ChartInstruct-LLaMA2 and ChartGemma. We also provide the p-values by performing Mann-Whitney U Tests.
    }
    \label{tab:evaluation-table-human}
\end{table}

We show the results of human evaluation when measuring the informativeness, factual correctness, and structure of outputs generated by ChartInstruct-LLaMA2 and ChartGemma on the 'Web' set of charts scraped from the web in \cref{tab:evaluation-table-human}. We see that ChartGemma significantly outperforms ChartInstruct-LLaMA2 in terms of informativeness and factual correctness and they match in the structure of the generated summary.

\subsection{Error Analysis}
\begin{figure*}[t!]
     \centering
        \scalebox{0.95}{\includegraphics[width=1\textwidth]{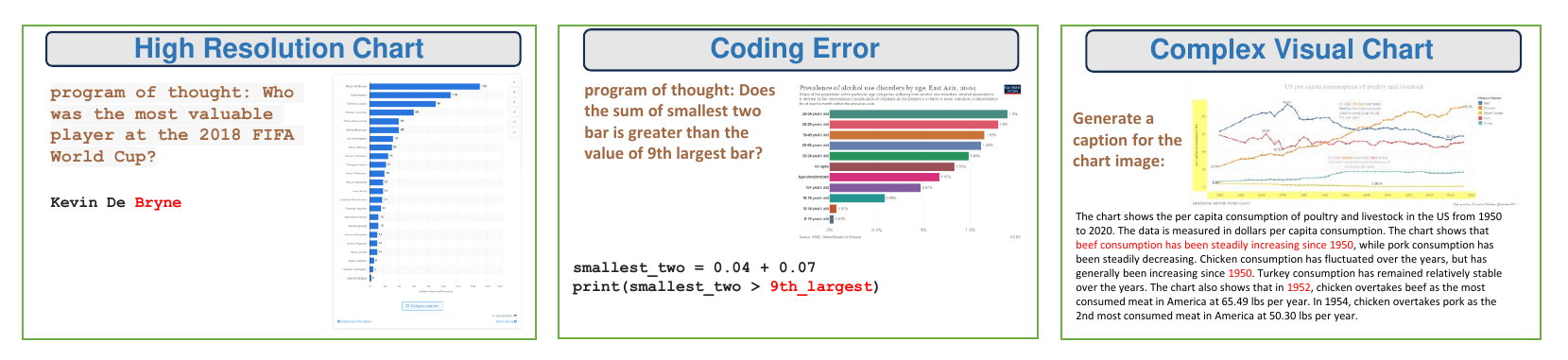}}
        \vspace{-3mm}
         \caption{\small{
         Some samples that our model, ChartGemma, has typos, coding errors, and factually incorrect statements in their outputs. The errors are shown in \red{red}. 
         }
         }\label{fig:WrongSamples}
\end{figure*}

\cref{fig:WrongSamples} show typoes and coding errors produced by our model.

\subsection{Sample Outputs from ChartGemma}
\begin{figure*}[t!]
    \centering
    \scalebox{0.95}{\includegraphics[width=1\textwidth]{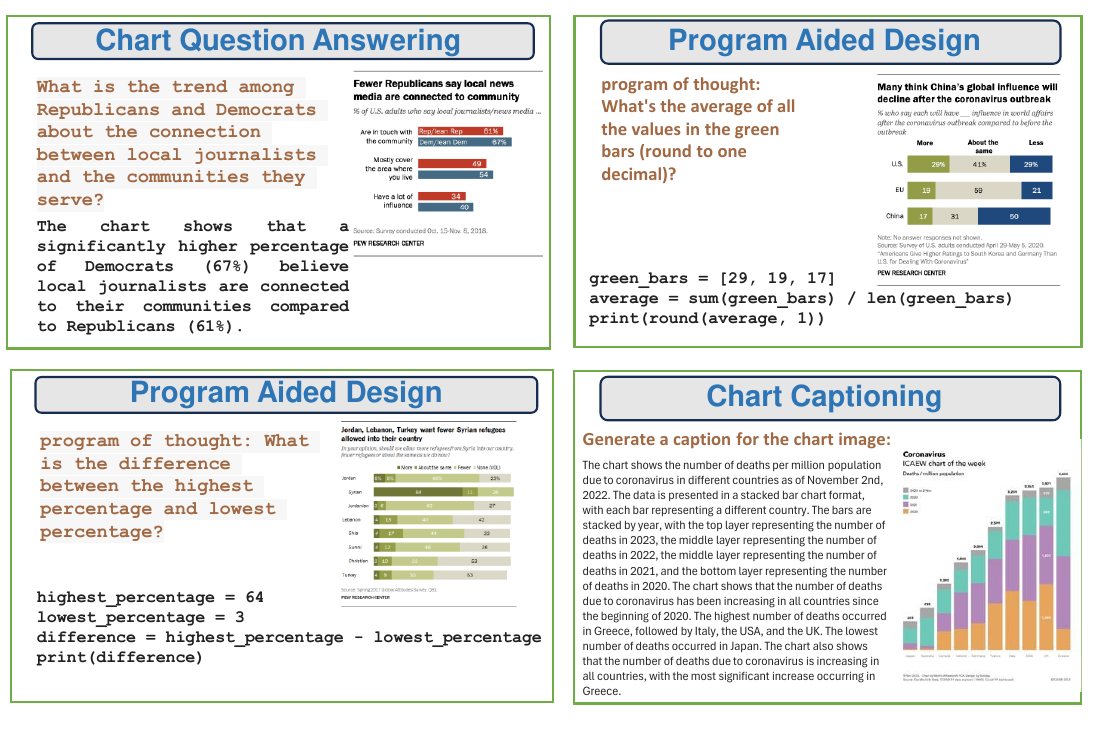}}
    \caption{Sample outputs generated by \model\ on various downstream tasks.}
    \label{fig:goods}
\end{figure*}

In \cref{fig:goods}, we provide some sample outputs on various tasks.

\end{appendices}

\end{document}